\documentclass[lettersize,journal]{IEEEtran}

\usepackage{tikz}
\usepackage{amsmath,amsfonts}
\usepackage{algorithmic}
\usepackage{algorithm}
\usepackage{array}
\usepackage[caption=false,font=normalsize,labelfont=sf,textfont=sf]{subfig}
\usepackage{textcomp}
\usepackage{stfloats}
\usepackage{url}
\usepackage{verbatim}
\usepackage{graphicx}
\usepackage{cite}
\usepackage{algorithm}
\usepackage{algorithmic}

\usepackage{makecell}
\usepackage{multirow}
\usepackage{newfloat}
\usepackage{listings}
\usepackage{xcolor}
\usepackage{xspace}
\usepackage{amsfonts}
\usepackage[show]{chato-notes}
\usepackage{adjustbox}
\usepackage{array}
\usepackage{booktabs}
\usepackage{bbm}
\usepackage{dsfont}
\usepackage{xspace}
\usepackage{url}
\usepackage{color,soul}
\usepackage{hyperref}

\newcommand{\cifar}{CIFAR-10\xspace}

\newcommand{\imagenet}{ImageNet\xspace}

\newcommand{\apb}{\texttt{APB}\xspace}

\newcommand{\fma}{\texttt{fma}\xspace}

\DeclareMathSymbol{\shortminus}{\mathbin}{AMSa}{"39}  

\usepackage{mathtools}

\begin{document}

\title{Neural Network Compression using Binarization\\and Few Full-Precision Weights}

\author{Franco Maria Nardini, Cosimo Rulli, Salvatore Trani, and Rossano Venturini
\IEEEcompsocitemizethanks{
\IEEEcompsocthanksitem{Franco Maria Nardini, Cosimo Rulli, and Salvatore Trani are with the National Research Council of Italy, Pisa, Italy. E-mail: \{\href{mailto:francomaria.nardini@isti.cnr.it}{francomaria.nardini}, \href{mailto:cosimo.rulli@isti.cnr.it}{cosimo.rulli}, \href{mailto:salvatore.trani@isti.cnr.it}{salvatore.trani}\}@isti.cnr.it}
\IEEEcompsocthanksitem{Rossano Venturini is with the Department of Computer Science, University of Pisa, Italy. E-mail:\href{mailto:rossano.venturini@unipi.it}{rossano.venturini@unipi.it}\vspace{2mm}}
}

\thanks{\textbf{Under Review at IEEE TKDE. September 2023.}}
}




\maketitle

\begin{abstract}
Quantization and pruning are two effective Deep Neural Networks model compression methods.
In this paper, we propose \emph{Automatic Prune Binarization} (\apb), a novel compression technique combining quantization with pruning. \apb enhances the representational capability of binary networks using a few full-precision weights.
Our technique jointly maximizes the accuracy of the network while minimizing its memory impact by deciding whether each weight should be binarized or kept in full precision.
We show how to efficiently perform a forward pass through layers compressed using \apb by decomposing it into a binary and a sparse-dense matrix multiplication. 
Moreover, we design two novel efficient algorithms for extremely quantized matrix multiplication on CPU, leveraging highly efficient bitwise operations. The proposed algorithms are $6.9\times$ and $1.5\times$ faster than available state-of-the-art solutions. 
We extensively evaluate \apb on two widely adopted model compression datasets, namely \cifar and \imagenet. \apb shows to deliver better accuracy/memory trade-off compared to state-of-the-art methods based on i) quantization, ii) pruning, and iii) a combination of pruning and quantization.
\apb also outperforms quantization in the accuracy/efficiency trade-off, being up to $2\times$ faster than the $2$-bits quantized model with no loss in accuracy.
\end{abstract}



\section{Introduction}
\label{sec:intro}
\IEEEPARstart{D}eep Neural Networks (DNNs) had an unprecedented impact on computer vision and achieve state-of-the-art performance in many different tasks, such as image classification~\cite{tan2019efficientnet}, semantic segmentation~\cite{he2017mask}, and object detection~\cite{redmon2016you}. Indeed, DNNs huge computational requirements pose severe challenges to their pervasive application. 
Model compression is the field of Deep Learning devoted to decreasing the computational requirements of Deep Neural Networks (DNNs) by leveraging their largely proven over-parametrization~\cite{bubeck2021universal}.

Quantization techniques reduce the number of bits required to represent the network parameters, thus offering compelling properties in terms of space saving and inference speedup. In this line, binarization, i.e., $1$-bit quantization, reduces the memory burden of $32\times$ with respect to its full-precision counterpart and converts floating-point multiplications into cheap bitwise operations~\cite{DBLP:journals/corr/RastegariORF16}.
Despite the work done in this field~\cite{qin2020forward,Xu_2021_ICCV,yang2020searching}, binary networks still struggle to match the performance of the corresponding full-precision model. At the same time, $2$ or $3$-bits quantization approaches are closing the gap with full precision models~\cite{lee2021network,yamamoto2021learnable}, but do not ensure the compelling properties of binary networks, especially in terms of inference speedup on CPU.
\begin{figure}[t!]
\centering
\includegraphics[width=1\columnwidth]{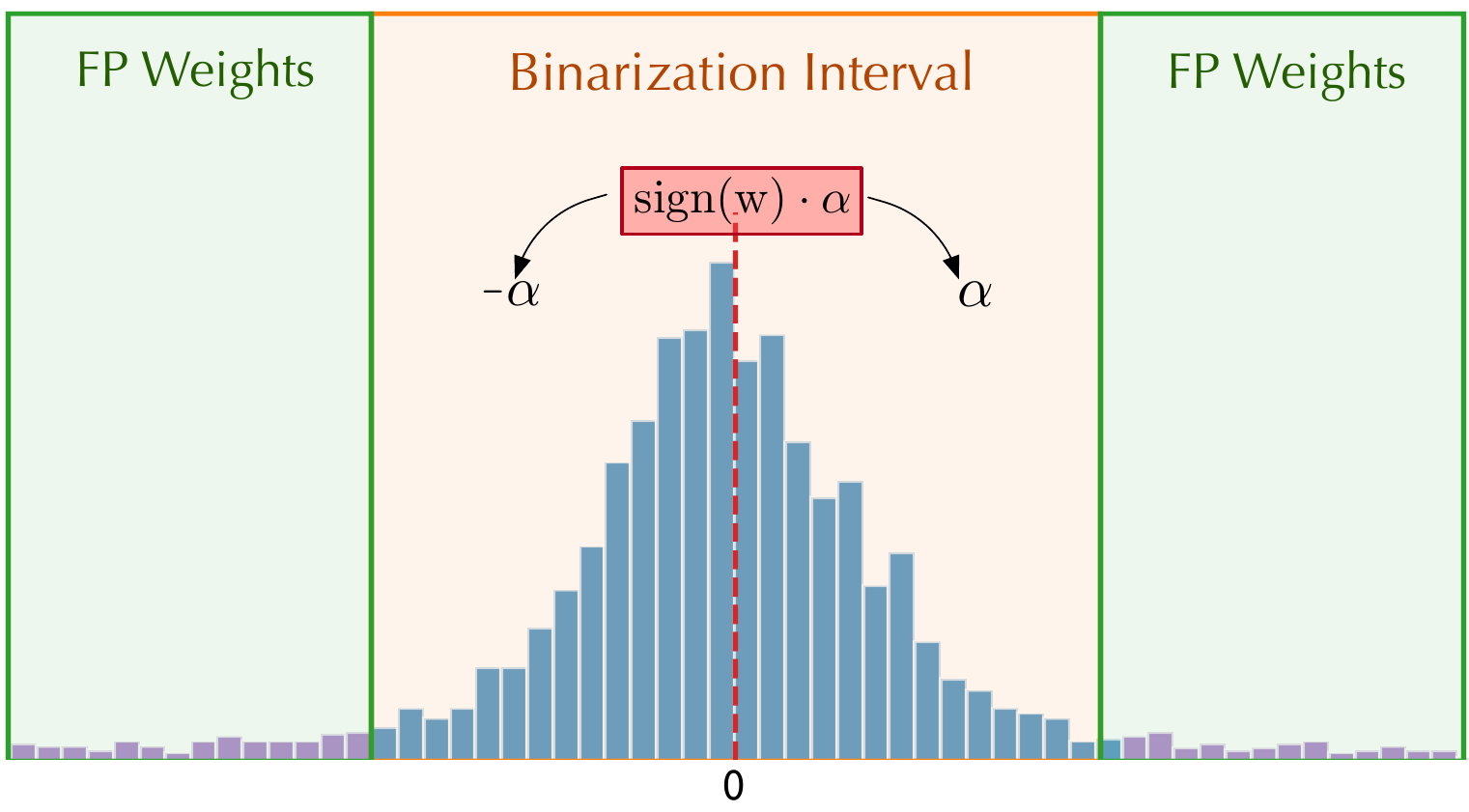}
\caption{Graphical illustration of Automatic Prune Binarization (\apb). The weights in the binarization interval are converted to $\{-\alpha, \alpha \}$ according to their sign while the remaining values are kept in full precision.\label{fig:intro_pic}}
\end{figure}

Pruning techniques remove network parameters to produce sparse models as accurate as their original dense versions~\cite{DBLP:journals/corr/HanPTD15,DBLP:journals/corr/GuoYC16,sanh2020movement}.
Pruning allows for consistent memory savings as only the non-zero parameters and their positions need to be saved. 
Despite the reduction of Floating Points OPerations (FLOPs), 
pruning a neural network does not imply a remarkable inference speedup until extreme sparsity levels are achieved, i.e., $> 95\%$, as CPUs and GPUs are optimized for dense computation~\cite{yu2017scalpel}. 
To summarize, binarization techniques allow for fast neural inference but do not match the performance of the equivalent full-precision models. On the other hand, pruning techniques can produce highly effective sparse networks but do not deliver consistent speedup until extreme sparsity ratios are reached.
Observe that quantization and pruning adopt complementary approaches to compress a neural network. Quantization maintains network parameters' cardinality while shrinking the parameters' representation, i.e., the number of distinct values a parameter can assume. Instead, pruning reduces the cardinality of non-trivial parameters, but it allows the surviving ones to span in the original representation, i.e., real numbers represented using $32$ bits. We aim to blend these approaches to leverage quantization properties, especially binarization, while empowering its representational capability with few full-precision entries.

In this paper, we propose Automatic Prune Binarization (\apb), a novel compression method that combines low-bit quantization (binarization) with pruning. \apb allows each parameter to assume either a binary or a full-precision representation (Figure~\ref{fig:intro_pic}).
\apb jointly maximizes the accuracy achieved by the network while minimizing its memory impact by identifying an optimal partition of the network parameters among these two sets.
In detail, it works by identifying a binarization interval (centered in~$0$), and the parameters falling in this interval are represented using one bit. As done by other state-of-the-art binarization approaches, according to their sign, binarized parameters are converted to $\{\shortminus \alpha, \alpha \}$, where $\alpha$ is a learned layer-wise scalar value (Figure~\ref{fig:intro_pic}). On the other hand, parameters outside the binarization interval are kept in full precision.
By doing so, \apb produces two different overlapping networks, i.e., a binary network and an extremely sparse full-precision network.
Their combined expressiveness allows for improving the performance of binary networks without the need to double the bits required to represent the weights, as done in $2$-bits quantization. 
This goal can be achieved when the number of full-precision weights is sufficiently low.
In this case, \apb also offers compelling inference properties.
In fact, we experimentally show that the overhead given by the sparse network is negligible at our sparsity ratios.
Moreover, we develop and present two novel matrix multiplication algorithms for CPU in extreme quantization scenarios. 
Our approach works by remapping a $q$-bits matrix in a set of $q+1$ binary matrices. Then, these matrices can be efficiently multiplied by leveraging cheap bitwise operations. 
We provide a high-performance implementation of these novel algorithms on CPU and 
show that our implementation can be much faster than currently available, general-purpose solutions~\cite{khudia2021fbgemm} for quantized matrix multiplication.
To the best of our knowledge, these are the first techniques tailored for extremely quantized matrices on CPU, except for the binary case. This allows the evaluation of the efficiency of highly quantized networks on CPU for the first time.
Overall, we experimentally show that \apb outperforms the performance of state-of-the-art quantization approaches in terms of memory/accuracy and efficiency/accuracy trade-offs. 

In detail, the novel contributions of this work are:
\begin{itemize}

    \item we introduce Automatic Prune Binarization (\apb). This novel compression framework jointly maximizes the accuracy achieved by the network while minimizing its memory impact by deciding whether each weight should be binarized or kept in full precision. We show how the problem of partitioning the set of weights can be addressed using Stochastic Gradient Descent.

    \item we address the problem of the efficiency of quantized networks on CPU. We first show that the forward pass through a layer compressed using \apb can be decomposed into a binary multiplication and a sparse binary multiplication.
    We discuss the efficiency of matrix multiplication as a function of the operand's bit width. We then design two novel matrix multiplication routines based on efficient bitwise operations available on CPU for extreme quantization scenarios. The source code of \apb and our novel bitwise matrix multiplication routines will be released upon publication of the manuscript.

    \item we provide an extensive experimental evaluation on two public datasets used in model compression, i.e., \cifar~\cite{cifar10} and \imagenet~\cite{imagenet_cvpr09}. Experiments show that \apb offers better accuracy/memory trade-off compared to any state-of-the-art compression method based on i) quantization, ii) pruning, and iii) the combination of quantization and pruning.

    \item we evaluate the performance of our novel low-bits matrix multiplication routines when employed for the forward pass of neural networks on 
    \imagenet. Our methods are $6.85\times$ and $1.5\times$ faster than available state-of-the-art solutions.  
    Moreover, \apb shows superior performance even in the accuracy/efficiency trade-off, being $2\times$ faster than the $2$-bits quantized model with no loss in accuracy.
 \end{itemize}

The rest of the paper is organized as follows: in Section~\ref{sec:related} we present the related work in binarization, low-bits quantization, pruning, and efficient inference on CPU for neural networks. In Section~\ref{sec:automatic_prune_binarization} we present our novel \apb compression framework mixing full-precision and binary weights. In Section~\ref{sec:bitwise} we discuss the efficiency of matrix multiplication at different bit widths and we introduce our novel low-bit matrix multiplication routines.
Section~\ref{sec:exp} presents an experimental evaluation of \apb. First, we discuss the memory compression performance of \apb (Section~\ref{subsec:comprperf}). Then, we evaluate the performance of our matrix multiplication algorithms and compare the execution time of the quantized networks against \apb~(Section~\ref{subsec:effeval}).
Finally, Section~\ref{sec:concl} concludes the work and draws some future lines of research.


\section{Related Work}
\label{sec:related}
Our method lies at the intersection of two families of compressors, i.e., \emph{binarization}/\emph{low-bit quantization}~\cite{courbariaux2015binaryconnect} and \emph{pruning}~\cite{lecun1990optimal,hassibi1993second}. In the following, we review the main contributions in these lines along with the ones investigating \emph{efficient inference} algorithms on quantized networks \cite{nurvitadhi2016accelerating}. 

\smallskip
\noindent \textbf{Binarization}.
Pioneering work on binarization are BinaryConnect~\cite{courbariaux2015binaryconnect} and XNOR-Net \cite{DBLP:journals/corr/RastegariORF16}. These methods rely on the use of the \emph{sign} function to constrain the weights in $\{\shortminus 1, 1 \}$ and to scale them by the mean of their absolute value.
More recent work leverages advanced techniques to train highly accurate binary models. 
Recently, in that line, some works show that maximizing the entropy of the binarized weights is an effective approach~\cite{qin2020forward,li2022equal}.
Xu \emph{et al.} show the importance of \emph{latent weights}, i.e.,  full-precision weights used during backpropagation and weight update. The authors focus on \emph{dead weights}, i.e., weights that are rarely updated due to their distance to the origin. Authors show that these weights are responsible for hampering the training process and they propose a tailored Rectified Clamp Unit to revivify those weights.
Liu \emph{et al.} tackle the problem of frequent weight flipping, i.e., weights changing their sign, by employing two learnable scaling gradient factors for the activations, one for each of the binary states ($\{\shortminus 1, 1\}$) \cite{liu2021sa}. 
Another family of approaches proposes architectural changes to the network to quantize. 
Bi-Real Net adds a double skip-connection on the ResNet architecture to sum the real-valued input with the features obtained after a binary convolution \cite{liu2018bi}.
A well-known solution in this field is ReActNet~\cite{liu2020reactnet}. Here, the authors duplicate the input channels of convolutional layers, introduce tailored activation functions for binary networks, and employ knowledge distillation to enhance the training phase. 
Hu \emph{et al.} introduce \emph{Squeeze} and \emph{Expand} layers aiming at combining input and output activations \cite{hu2022elastic}.
For a comprehensive discussion on binary networks, we recommend BiBench~\cite{DBLP:conf/icml/QinZDLC0Y023}, a recently released benchmark comparing the performance of binarization algorithms on different tasks and different architectures. We remind the reader that this work only focuses on binarization algorithms for convolutional neural networks.

\smallskip
\noindent \textbf{Low-bits Quantization}.
Quantization techniques often rely on the Straight-Through Estimator (STE)~\cite{bengio2013estimating} to propagate the gradients through non-differentiable quantization functions.
Lee \emph{et al.} propose to overcome the limits of STE by exploiting an element-wise gradient correction method~\cite{lee2021network}. Their approach, named Element-Wise Gradient Scaling (EWGS), scales the gradient of each full-precision weight according to three factors: i) its sign, ii) the gap between full-precision and quantized value, iii) a scaling factor learned through the approximation of the Hessian.
SLB employs a continuous relaxation strategy to overcome the gradient mismatch problem~\cite{yang2020searching}. Each weight of the network is mapped to a probability distribution that represents the values it can assume with a $q$-bits representation.
The values associated with the highest probabilities are selected as quantization values.
Yamamoto proposes a non-uniform quantization method for pre-activations ResNet models~\cite{he2016identity}. The approach is based on learnable functions that wrap the quantization process to allow effective tuning of the quantization levels~\cite{yamamoto2021learnable}.

\smallskip
\noindent \textbf{Pruning}.
Pruning techniques effectively sparsify neural networks with small/no accuracy degradation.\footnote{with pruning, we always refer to element-wise pruning. See Liang \emph{et al.}~\cite{liang2021pruning} for a complete analysis of the difference between element-wise and structured pruning. }
Recently, a plethora of different pruning methods has been developed. For a comprehensive discussion, we recommend the reading of dedicated surveys~\cite{liang2021pruning}. We highlight that
the importance of high absolute-value weights in neural networks was first discovered in pruning techniques.  In fact, magnitude-based heuristics save high absolute-value weights while zeroing out the others. Han \emph{et al.} are the first to apply magnitude-based pruning in conjunction with re-training of the network to mitigate the possible performance degradation~\cite{DBLP:journals/corr/HanMD15}.
Lately, magnitude pruning has been improved by introducing re-winding, namely reassigning the surviving parameters to their initialization values, showing that it outperforms fine-tuning on several network architectures and datasets~\cite{renda2019comparing}. A lot of effort has been spent in training sparse networks (or pruning them at initialization), rather than pruning after training. This interest is motivated by the so-called ``Lottery Ticket Hypothesis'', namely the existence of highly effective sparse networks in randomly initialized dense networks~\cite{frankle2018lottery}.

\smallskip
\noindent \textbf{Combination of Pruning and Quantization}
Several works explore the combination of pruning and quantization, to fruitfully exploit the advantages provided by both techniques~\cite{DBLP:journals/corr/HanMD15,tung2018clip,van2020bayesian,diffenderfer2020multi,li2022equal}. Han \emph{et al.}~\cite{DBLP:journals/corr/HanMD15} apply clustering algorithm on the weights surviving the pruning phase, then optimize the value of the centroids using the average of the weight gradients. 
Bayesian Bits~\cite{van2020bayesian} is an approach where mixed precision quantization is combined with pruning. In particular, network weights are either quantized to a power-of-two-bit width or zeroed out in a data-driven fashion. In ``Multi-Prize Lottery Ticket'' (MPT)~\cite{diffenderfer2020multi}, Diffenderfer \emph{et al.} mix pruning and binarization by i) discovering highly-effective sub-networks in neural networks, ii) binarizing the surviving values.
Recently, a combination of pruning and quantization approaches have been applied to transformer-based architectures. In this line, SPQR~\cite{dettmers2023spqr} isolates outlier weights in each layer and keeps them in full precision while quantizing the remaining ones using $3$ or $4$ bit per weight.
SPQR differs from \apb for two main reasons: first, SPQR isolates \emph{blocks} of weights, such as rows or columns, while \apb works on single tensor elements. Second, SPQR is a post-training quantization method, while \apb leverages training-time gradients to partition binary and full-precision weights optimally.\footnote{Our approach was developed simultaneously and independently from SPQR.}

\smallskip
\noindent \textbf{Efficient Inference}.
Nurvitadhi \emph{et al}. study the efficiency of binary multiplication on different hardware platforms. Authors estimate a speedup of $2\times$ of binary over single-precision multiplication on a CPU equipped with $64$-bit bitwise instructions \cite{nurvitadhi2016accelerating}.

Regarding efficient inference of binary network on CPU,  a major contribution is BitFlow, a  binary convolution algorithm (\emph{Pressed-Conv}) based on bit-packing on the channel dimension~\cite{hu2018bitflow}. BitFlow provides $1.8\times$ speedup compared to na\"ive binary convolution. In this line, DaBnn~\cite{zhang2019dabnn} and Larq~\cite{geiger2020open} are efficient inference frameworks for binary neural networks on mobile platforms powered by ARM processors. 
Our work is the first to study the efficiency of low-bit quantized neural networks on general-purpose CPUs.

\smallskip
\noindent \textbf{Our Contribution}.
\apb approach is orthogonal to other techniques mixing pruning and quantization.
As an example, in works combining binarization and pruning~\cite{diffenderfer2020multi}, parameters are either zero or binary ($
\{ \shortminus 1, +1\})$. 
\apb, instead, enriches the expressiveness of binary networks with full-precision parameters in the same framework that effectively mixes binarization and pruning. 
This means that weights in \apb-compressed networks are either binary or \emph{full-precision}.
To the best of our knowledge, we are the first to jointly optimize binary and full-precision parameters in a novel and end-to-end compression framework.


\section{Automatic Prune Binarization}
\label{sec:automatic_prune_binarization}

We now describe \apb, our novel compression framework that adds a few full-precision values to binary networks. First, we introduce the role of binarization. Second, we show how large absolute-value weights play different roles in binarization and pruning. Third, we formally introduce \apb and show how its parameters can be optimized by leveraging Stochastic Gradient Descent (SGD). Finally, we show how to decompose the matrix multiplication into dense-dense and sparse-dense matrix multiplications. 

\smallskip
\noindent \textbf{Binarization}.
Let us consider $W \in \mathbb{R}^n$ as the set of weights of a neural network. The scope of binarization is to employ a single bit to store each weight $w \in W$, forcing $w$ to be in $\{\shortminus 1, +1\}$. 
Previous studies show that it is possible to enhance the expressiveness of the model by scaling the weights with a scalar $\alpha$~\cite{DBLP:journals/corr/RastegariORF16} so as to remap them to $\{\shortminus \alpha, +\alpha \}$.
We rely on re-scaled binarization that asks for the definition of a $\mathtt{Bin}$ operator defined as follows:

\begin{equation}
\label{eq:binariation}
\mathtt{Bin}(w)= \alpha \cdot \text{sign}(w)  = \left\{\begin{array}{ll}
+ \alpha & \text { if } w \geq 0\\
 - \alpha & \text { if } w < 0.
\end{array}\right.
\end{equation}

The $\mathtt{Bin}$ operator is defined as a function of $sign(w)$, whose derivative is zero almost everywhere. This hinders the usage of gradient-based approaches for training the model.
For this reason, gradients are approximated with the Straight Through Estimator (STE) \cite{bengio2013estimating}.
STE works by using a surrogate differentiable function $g(w)$ that approximates $sign(w)$. 
Hence, we can use the derivative of $g()$ in place of the derivative of $sign()$.
In practice, STE imposes that:
\begin{equation}
\label{eq:ste}
\frac{\partial \, \mathtt{Bin(w)}}{\partial w } = \frac{\partial g(w)}{\partial w }.
\end{equation}

This derivative is used to update the full-precision weights $W_l$, which are referred as \emph{latent weights}~\cite{Xu_2021_ICCV} in the context of binarization. The final binary matrix $W$ is then obtained by simply applying the $\mathtt{Bin}$ operator over $W_l$.

\smallskip
\noindent \textbf{Large Absolute-value Weights}.
Given the STE training strategy, the binary weight $w$ is updated only when there is a flip of sign in the corresponding latent weight $w_l \in W_l$, as a result of the gradient update.
In a recent work~\cite{Xu_2021_ICCV}, authors identified the problem of large absolute-value parameters that diverge from the zero-centered Laplacian distribution characterizing latent weights, namely \emph{dead weights}. 
These weights interfere with the optimization phase giving a reduced chance to change their sign. 
In practice, they freeze part of the network, hindering the training process.

\begin{figure}[t]
  \centering
   \includegraphics[width=0.9\columnwidth]{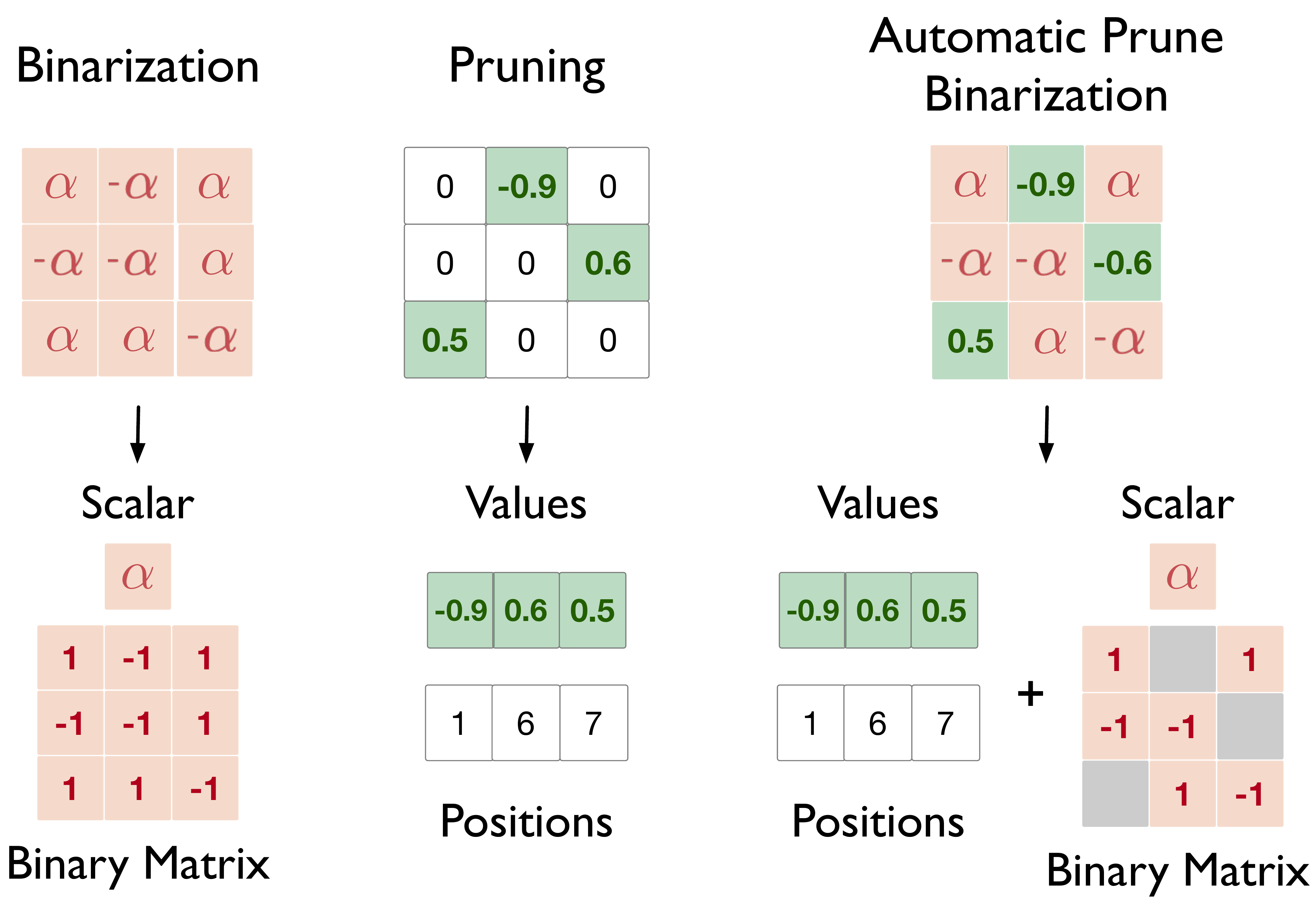}
   \caption{Network weights representation in binarization (left), pruning (center), \apb (right). In \apb, binary and full-precision weights coexist in the same matrix. For full-precision entries, we require to store also the index inside the weight matrix.}
   \label{fig:prunebin_2}
\end{figure}

Even in pruning, large absolute-value weights play a central role. Here,  network layers are sparsified by zeroing out less important weights, and the importance of each parameter is chosen according to a heuristic. Interestingly, a simple yet very effective heuristic to determine the importance of a parameter is its magnitude~\cite{DBLP:journals/corr/HanPTD15,DBLP:journals/corr/GuoYC16}. 
Several works show that a small portion ($< 10\%$) of large absolute-value weights is enough to match the performance of the dense model~\cite{DBLP:journals/corr/HanPTD15,DBLP:journals/corr/GuoYC16,gale2019state}.
Large absolute-value weights thus play contrasting roles in binarization and pruning techniques. In binarization, they hamper and slow down the training process due to the low likelihood of changing the sign. In pruning, they synthesize the expressiveness of the overall model. We build our approach on this discrepancy: we leverage the advantages that large-absolute weights provide in pruning while mitigating the drawbacks associated with binarization. 
This is the key intuition underlying \apb, our novel method for effective compression of neural networks. \apb keeps high absolute-value weights in full precision, and it binarizes the remaining ones. 
For this purpose, \apb defines a symmetric binarization interval around zero (Figure~\ref{fig:intro_pic}): weights falling outside this interval are kept in full-precision, while the others are mapped to $\{\shortminus \alpha, + \alpha\}$ according to their sign. In this regard, \apb can be interpreted as a pruning technique where small absolute-value weights are binarized instead of being zeroed out~\cite{DBLP:journals/corr/HanPTD15}. The amplitude of the binarization interval and the value of the scalar $\alpha$ are optimized during training.
As shown in Figure~\ref{fig:prunebin_2}, \apb compresses the network by applying pruning and binarization in parallel, i.e., each weight is either full-precision or binary. 
Conversely, in classical approaches mixing pruning and quantization/binarization, each parameter is either zeroed-out or quantized/binarized~\cite{DBLP:journals/corr/HanMD15,tung2018clip,van2020bayesian}.
Within \apb, two different networks coexist during the training: a binary and a sparse network. 

\begin{figure}[t]
  \centering
   \includegraphics[width=\columnwidth]{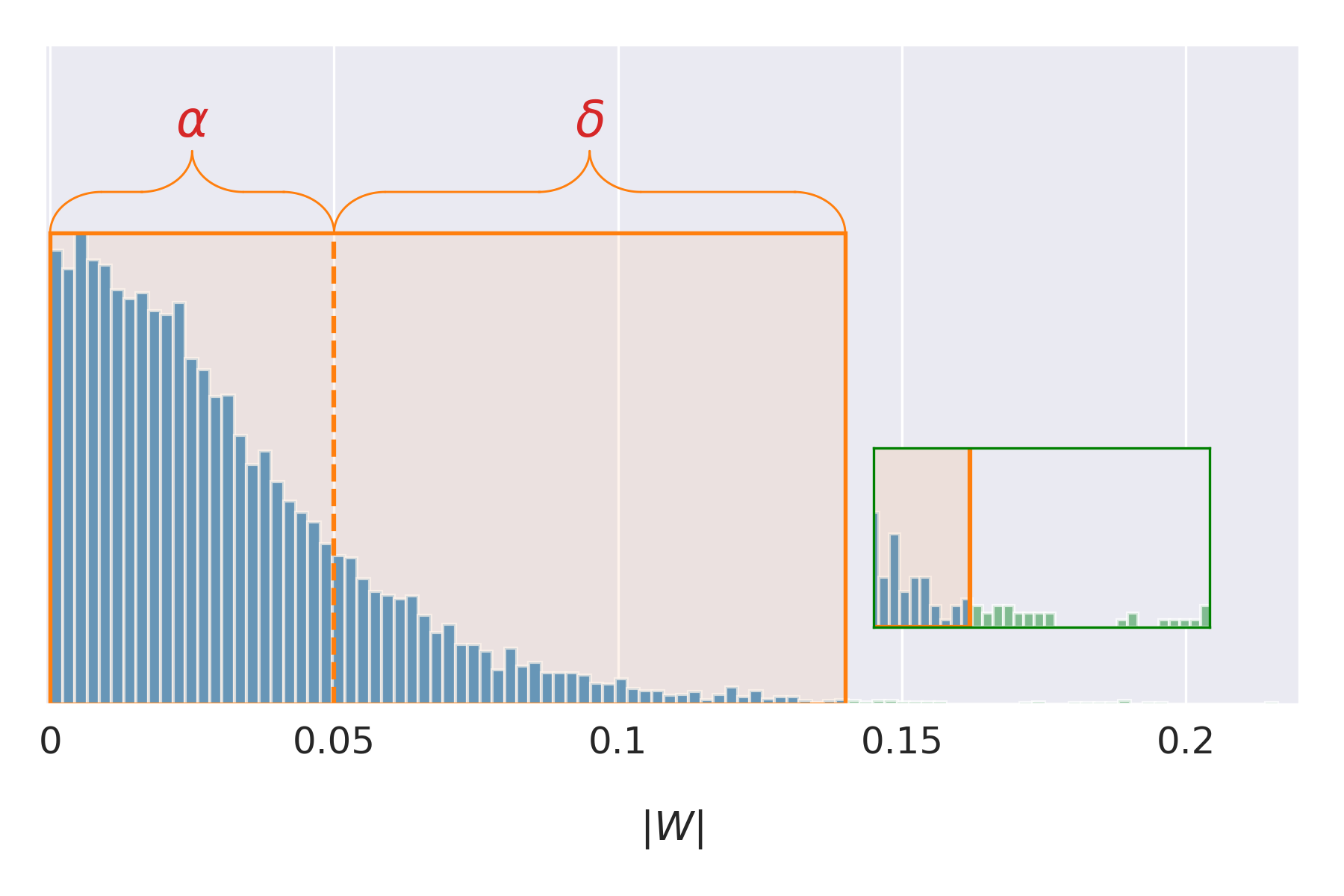}
   \vspace{-10mm}
   \caption{Automatic Prune Binarization ($\mathtt{APB}$) applied on the network parameters $W$. The width of the binarization interval (orange area) is defined by $\alpha + \delta$. A weight $w_i \in W$ is binarized if $|w_i| \leq \alpha + \delta$, otherwise it is kept in full-precision (green area).}
   \label{fig:zoom_prunebin}
   \vspace{-2mm}
\end{figure}

\noindent \textbf{\apb}.
Given the considerations on the role of large absolute-value weights, \apb employs weight magnitude to determine whether weight should be binarized or kept in full precision.
Our approach is partially inspired by~\cite{wang2020sparsity}, which introduces a trainable threshold to determine if a weight should be set to $0$ or $1$.
The operator \apb on a weight $w$ is defined as:
\begin{equation}
\label{eq:APB}
\mathtt{APB(w)}=\left\{\begin{array}{ll}
\text{sign}(w) \,  \alpha & \text { if }  |w| \leq \alpha + \delta \\
w& \text { otherwise, }
\end{array}\right.
\end{equation}
with $\delta$ being the amplitude of the binarization interval exceeding $\alpha$. 
Figure~\ref{fig:zoom_prunebin} graphically depicts how \apb works.
The weights whose absolute value ranges in $[0, \alpha + \delta] $ are binarized (orange area), while the parameters falling outside this interval are kept in full precision (green area). 

If $w$ is within the binarization interval, \apb is not differentiable. In this case, we apply STE
by employing the identity function, i.e., $id(x) = x$, as $g(w)$~\cite{bengio2013estimating}. Thus, the derivative becomes:
    
\begin{equation}
    \label{eq:STE_apb}
\frac{\partial \mathtt{APB}}{\partial w}=\left\{\begin{array}{ll}
g'(w) & \text { if } |w| \leq \alpha + \delta \\
1 & \text { otherwise. }
\end{array}\right.
\end{equation}
Assuming $\delta \geq 0$, we can re-write

\begin{equation}
\label{eq:absval}
|w| \leq \alpha + \delta \quad \Rightarrow \quad \frac{|w| - \alpha}{ \delta} \leq 1. 
\end{equation}
To ease the notation, we define $\hat{w} = \frac{|w| - \alpha}{ \delta}$. This entails: 
\begin{equation}
\label{eq:apb_hat}
\mathcal{\mathtt{APB}}(w)=\left\{\begin{array}{ll}
\text{sign}(w) \, \alpha & \text { if }  \hat{w} \leq 1 \\
w& \text { otherwise. }
\end{array}\right.
\end{equation}

We define the indicator function of the set of binarized weight as $\chi_B : = 
\mathds{1}
(\hat{w}\leq 1)$.
 We can now define the gradients of $\alpha$ and $\delta$ by unrolling the derivatives of the loss function $\mathcal{L}$. We introduce the indicator function to constrain $\alpha$ and $\delta$ to depend exclusively on the binarized weights and by leaving them independent from those weights that \apb lefts full-precision. 


\begin{equation}
\label{eq:der_delta}
\frac{\partial \mathcal{L}}{\partial \delta} = \frac{1}{n} \sum   \frac{\partial \mathcal{L}}{\partial \hat{w} }  \frac{\partial \hat{w}}{\partial \delta } \chi_B = \frac{1}{ \delta^2 n }  \sum \frac{\partial\mathcal{L}  }{\partial \hat{w} }(\alpha - |w|)\chi_B.
\end{equation}


\begin{equation}
\label{eq:der_alpha}
\frac{\partial \mathcal{L}}{\partial \alpha} =  \frac{1}{n} \sum \frac{\partial \mathcal{L}}{\partial \hat{w}}
\frac{\partial \hat{w}}{\partial \alpha} =
- \frac{1}{\delta n} \sum \frac{\partial\mathcal{L}  }{\partial \hat{w} } \chi_B.
\end{equation}

We also observe that
\begin{equation}
    \frac{\partial \mathcal{L}}{\partial w}  = \frac{1}{\delta} \frac{\partial \mathcal{L}}{\partial \hat{w}} \, \text{sign}(w).   
\end{equation}

We can compute $\frac{\partial \mathcal{L}}{\partial \alpha}$ and $\frac{\partial \mathcal{L}}{\partial \delta}$ using $\frac{\partial \mathcal{L}}{\partial w}$, which is the standard derivative of the weights w.r.t to the cost function, obtained using the backpropagation algorithm~\cite{rumelhart1986learning}. 




\noindent \textbf{Memory Impact}. 
We denote with $\mathtt{Mem}(W)$ the memory impact of the matrix $W \in \mathbb{R}^n$ compressed using \apb, expressed in bits.
Assuming to have $s$ full-precision surviving entries, we need to store $n-s$ binary weights and $s$ full-precision values with their position. However, since $s \ll n$ and for easing the matrix multiplication, the binary matrix is fully represented.
Hence:
\begin{equation}
    \label{eq:bm}
    \mathtt{Mem}(W) = (n-s) + s(b_v + b_p) \simeq n + s(b_v + b_p)
\end{equation}
where $b_v$ is the bit width of the full precision values ($32$ for floating point) and  $b_p$ represents the bits needed to store their positions in the matrix. For each neural architecture under evaluation, we 
compute $b_p$ as $\max_{i} \log_2(k_i-1)+1$, where $k_i$ is the dimension of layer $i$.

\noindent \textbf{Inference Considerations}.
We now discuss how to efficiently multiply a weight matrix $A$, compressed using \apb, against an input activation matrix $B$.\footnote{Convolution can be converted to matrix multiplication using the \emph{im2col} technique~\cite{chellapilla2006high}.}
For every $A_i \in A$ --- where $A_i$ is an element of the input matrix --- we define the mask associated with $A$:
\begin{equation}
\label{eq:mask}
\mathtt{Mask}(A_i)=\left\{\begin{array}{ll}
1 & \text { if } A_i \in \{ \shortminus\alpha, +\alpha \}\\
0 & \text { otherwise. }
\end{array}\right.
\end{equation}
%
We decompose $A =  A^{\text{bin}} + A^{\text{full}}$, where 
$A^{\text{bin}}  \in \{\shortminus\alpha,+\alpha\}^n $ is a binary matrix and $A^{\text{full}} \in \mathbb{R}^n $ is a sparse full-precision matrix.
$A^{\text{bin}}$ is defined as $A_i^{\text{bin}} =\alpha \cdot \text {sign}(A_i)$, while 
$A^{\text{full}}$ is given by
\begin{equation}
    \label{eq:asparse} 
    A^{\text{full}}_i = \left\{\begin{array}{ll}
     0& \text { if } \mathtt{Mask}(A_i) =1\\
    A_i - A^{\text{bin}}_i   & \text { otherwise. }
    \end{array}\right.
\end{equation}
Given an input matrix $B$ and the distributive property of matrix multiplication, we can write
\begin{equation}
    \label{eq:mmdistprop}
    C= A \cdot B = (A^{\text{bin}} + A^{\text{full}}) \cdot B = A^{\text{bin}} \cdot B +  A^{\text{full}} \cdot B.
\end{equation}
Since $A^\text{bin}$ is a binary matrix, the only overhead introduced by \apb is a sparse-dense matrix multiplication. Due to the extreme sparsity ratios of  $A^{\text{full}} $, the sparse-dense multiplication can be efficiently performed with tailored implementations such as LIBXSMM~\cite{heinecke2016libxsmm}.


\section{Bitwise Matrix Multiplication}
\label{sec:bitwise}
The efficiency of neural inference heavily relies on the efficiency of matrix multiplication (MM).
MM has been widely studied~\cite{goto2008anatomy,huang2016blislab,van2015blis} due to its paramount role in many scientific applications.
Here, we discuss the efficiency of MM at different quantization levels and we introduce our novel routines for optimized matrix multiplication in low-bit configurations.
In the following, we employ the notation $\mathtt{w} / \mathtt{a}$ to specify the bit width of the weights and the activations matrices, respectively.
In particular, we show how to implement efficient $1\text{/}2, 2\text{/}2$ MM using logical and bitwise operators. To the best of our knowledge, we are the first to investigate the efficiency of such low-bit configurations on CPU.

\noindent\textbf{Efficiency of Matrix Multiplication}.
Matrix multiplication is known to be a memory-bounded problem. Indeed, several techniques have been developed to mitigate this aspect, and nowadays, its efficiency mostly depends on the available computational power.\footnote{\url{https://en.algorithmica.org/hpc/algorithms/matmul/}}
The core operation of MM is the \emph{update} function $c \leftarrow c + ab$, which is recursively applied on portions of the input matrices $a,b$ to incrementally compute the output $c$. In the context of neural inference, $c$, $a$, and $b$ represent the output, the weights, and the input of each layer, respectively.
Modern CPUs allow performing the operation above with the \emph{fused-multiply add} (\fma) instruction, which computes the update with the same latency and throughput of an \texttt{add} instruction~\cite{fog2018instruction}. 
Furthermore, modern CPUs can rely on instruction-level parallelism (SIMD), which allows the processing of multiple inputs at the same time. As an example, the \texttt{\_mm512\_fmadd\_ps} instruction computes the operation $c \leftarrow c + a b$ on three vectors of $16$ full-precision values each.
The theoretical peak performance ($tpp$) measures how many update functions (up) can be carried out per second (sec) in an ideal scenario assuming that the memory cost is negligible, i.e., 
\begin{equation}
\label{eq:tpp}
tpp = cf \cdot tp \cdot v \  \frac{\text{up}}{\text{sec}},    
\end{equation}
where $cf$ is the clock frequency, $tp$ is the throughput of the \fma instruction and  $v$ is the number of operands that can be stored on a CPU register. $v$ is computed by dividing the width of the SIMD register, e.g., $512$ for \textsc{avx-512}, by the number of bits of the operand.

Network quantization exploits the performance gain obtained by reducing the bit width of weights and activations, hinging on the extreme flexibility of neural networks to model compression.
Observe that every time that the operand bit width halves, twice the data fit into the same CPU register ($v$). Consequently, $tpp$ doubles each time the operand bit width halves.
However, on modern CPUs the \texttt{fma} instruction exists only for double/single/half-precision float and $8$-bits integer values. For smaller bit widths, e.g., $2$ or $3$ bits, the use of the \texttt{fma} instruction requires to up-cast the operands to the closest supported data type, i.e., $8$ bits.
The situation is different for a binary network as it allows to implement MM by leveraging addition/subtraction or bitwise operations. This motivates us to investigate fast bitwise matrix multiplication techniques for efficient neural inference on CPU.
We now discuss the $1 \text{/} 32$ and the $1\text{/}1$ cases, then we present our novel bitwise matrix multiplication routines for the $1\text{/}2$ and the $2\text{/}2$ scenarios.


\smallskip
\noindent\textbf{1/32}.
In this quantization schema, weights $w$ are constrained to assume values in $\{\shortminus 1, 1  \}$. Conversely, activations $a$ are kept in full precision, i.e., they are represented by using 32-bit floating point.
This configuration is explored in several state-of-the-art quantization works~\cite{qin2020forward,gong2019differentiable,zhou2016dorefa}, as it features 
a $32\times$ memory saving compared to the single-precision representation. In principle, binary weights convert multiplication into additions and subtractions~\cite{DBLP:journals/corr/RastegariORF16}. 
We will now show that this conversion does not deliver remarkable speedup over classical multiplication.
Given $w \in \{\shortminus 1, 1\}$, in Equation~\ref{app:eq:1_32} we show how to rewrite the dot product between $w$ and $a$ into a series of additions and subtractions. We can write, 
\begin{equation}
    \label{app:eq:1_32}
    w \cdot  a =  \sum_{i=1}^n w_i a_i = \sum_{j \in \mathcal{I_{+}}} a_j - \sum_{j \in \mathcal{I_{-}}} a_j.
\end{equation}
where $ \mathcal{I_{+}} = \{ i \ | \ w_i =1 \}$ and $ \mathcal{I_{-}} = \{ i \ | \ w_i =\shortminus 1 \}$. 
In detail, $I_{+}$ is the set of indexes of positive weights ($w_i = 1$), while $I_{-}$ keeps track of negative weights ($w_i = \shortminus 1$). In this formulation, we simply accumulate the activations in correspondence with positive weights and then subtract the sum of the activations in correspondence with negative weights.
Indeed, the \fma operation achieves the same latency and throughput as the \texttt{add}/\texttt{sub} operations on modern CPUs.\footnote{\url{https://www.intel.com/content/www/us/en/docs/intrinsics-guide/index.html}}
This means that the update function $c \leftarrow c + ab$ (Equation~\ref{app:eq:1_32}) costs $2 \times$ the \fma-based full-precision matrix multiplication.
We can think of a more efficient approach that reduces the update function to a single addition operation.
Let us define
\begin{equation}
    \label{app:eq:1_32_aux}
    T = \sum_{i=1}^n a_i, \quad I_+ =  \sum_{j \in \mathcal{I_{+}}} a_j, \quad I_{-}=\sum_{j \in \mathcal{I_{-}}} a_j.
\end{equation}
with $T = I_{+} + I_{-}$ by construction.
$T$ is the sum of the activations along the columns and it does not depend on the weights $w$. This means that $T$ can be computed at the beginning of the multiplication and then re-used for all the columns of $a$. Moreover, it can be computed in $\Theta(n^2)$ time as it requires the sum of $n$ values along $n$ columns. Its impact is negligible compared to $\Theta(n^3)$ --- time needed to run the matrix multiplication.
$T$ can be exploited to avoid the computation of one between $I_{+}$ or $I_{-}$. For example, we can obtain $I_{-}$ as $I_{-} = 2\,I_{+}  - T $. Thus,
we can compute $w \cdot a$ as  $$w \cdot  a = I_{+} - I_{-} = 2\,I_{+}  - T.$$
The cost of multiplying by $2$ and subtracting $T$ is negligible, as it does not depend on the size of the input.
The cost of multiplying $w \in \{\shortminus 1, 1 \}$ and $a$ is dominated by the computation of $I_{+}$. It can be efficiently implemented by leveraging a masked floating point addition, where the mask identifies the $i \in I_{+}$. On recent Intel's instruction set AVX512, the masked floating point addition can be implemented by using the \texttt{\_mm512\_add\_ps} instruction. Anyway, this instruction has the same latency and throughput of \fma. As a consequence, it does not offer any computational advantage compared to 32-bit floating point MM.
Moreover, the binary representation for the weights is useless in this case, as they require to be converted into 32-bit float numbers when performing the vectorized  \texttt{\_mm512\_add\_ps} on CPU. This can be done in two ways: i) the representation of the binary weights is expanded in memory with the consequence of achieving the same memory footprint of full-precision computation, or ii) they are directly expanded to $32$ bits when moved to the CPU registers with the consequence of having an overhead due to bit extraction. We conclude that the $1\text{/}32$ scenario does not offer strong computational advantages compared to $32$ floating point implementation. 
The observations we draw also hold if activations are quantized to $8$ or $16$ bits.

\smallskip
\noindent\textbf{1/1}.
Several works investigate how to effectively train fully binary neural networks~\cite{DBLP:journals/corr/RastegariORF16,qin2020forward,Xu_2021_ICCV,liu2021sa,li2022equal,hu2022elastic}, where both weights $w$ and activations $a$ can assume two values, i.e., $\{\shortminus 1, +1\}$. Besides the memory savings achieved, $1/1$ also allows fast inference. In fact, 
given two binary vectors $u, v  \in \{ \shortminus 1, 1 \}^n$, their dot product can be computed as
\begin{equation}
\label{eq:binmult}
   u \cdot v = n -  2 \cdot \texttt{popcount}(\texttt{xor}(u, v)). 
\end{equation}
We observe that Equation~\ref{eq:binmult} holds for every $u^{\gamma} = \{\shortminus \gamma, \gamma \}^n$, $v^{\beta} = \{\shortminus \beta, \beta \}^n$, as $u^{\gamma} \cdot v^{\beta} = \gamma \beta  (u  \cdot v$). 
It is, in fact, a common practice to add scalar multipliers to enrich 
the expressiveness of binary networks.

We compute the $tpp$ for binary multiplication to compare it with full-precision computation. This requires estimating the optimal execution flow for the three \textsc{avx-512} operators (\texttt{xor, popcount, add}) required to perform the update function.
We assume to work on a modern CPU such as the Intel SkyLake architecture, equipped with $512$-bit registers and the \texttt{vpopcntq} instruction that enables the computation of the \texttt{popcount} operation on $512$-bit registers.  
Recall that modern architectures have different execution units, each associated with a different port.\footnote{\url{https://easyperf.net/blog/2018/03/21/port-contention}} Instructions employing different ports can be executed during the same clock cycle. Moreover, if an instruction uses $k$ different ports, it is executed $k$ times in the same clock cycle.
In this architecture, \texttt{xor} and \texttt{add} work on port $\{0, 5\}$ (throughput $2$), while \texttt{popcount} works exclusively on port $5$ (throughput $1$)~\cite{fog2018instruction}.
Hence, even in a perfectly pipelined execution flow, the computation of the update function in a single clock cycle, as done in the dense case using the \fma operation, is unfeasible. This would be possible only if all the operations (\texttt{xor}, \texttt{popcount}, and \texttt{add}) were assigned to different ports. In the best-case scenario, we can perform $2$ complete updates ($6$ instructions, $1024$ values) every $3$ clock cycle ($2$ instructions per clock cycle).
The number of values processed per clock cycle, on average, is$\frac{1024}{3} \simeq 341$. The same processor, equipped with  $\textsc{avx-512}$ registers and the \fma instruction with throughput $2$, can process $\frac{512}{32} = 16$ 32-bits floating point values per port, namely $32$ floating point values per clock cycle. This means that the theoretical speedup offered by binary multiplication is $10.5 \times$.
We show that our implementation of binary multiplication reaches the theoretical speedup compared to a state-of-the-art library for matrix multiplication, oneDNN.\footnote{\url{https://github.com/oneapi-src/oneDNN}}
In practice, the empirical speedup over the equivalent single-precision MM can be up to $16\times$ on rectangular matrices. In fact, it has been shown that dense multiplication does not reach the $tpp$ on non-squared matrices~\cite{nardini2022distilled}, while our bitwise routine can. We will detail this aspect in our experimental evaluation, Section~\ref{subsec:effeval}.

\smallskip
\noindent \textbf{1/2}.
This quantization schema achieves more accurate models than $1\text{/}1$.
State-of-the-art quantization approaches such as SLB~\cite{yang2020searching} and EWGS~\cite{lee2021network} have shown that $1\text{/}2$ delivers up to $9$ points of Top1 accuracy gain on the Image Classification task on \imagenet compared to the $1\text{/}1$ quantization.
Despite the effectiveness improvements achieved, no efficient MM routines have been proposed for this configuration on CPU.
In this regard, we design a novel matrix multiplication routine for $1\text{/}2$ quantization that exploits bitwise operations as in the $1\text{/}1$ case. 

Let us consider a binary vector $w \in \{ \shortminus \gamma, \gamma \}^n$ and a 2-bits uniformly quantized vector $a = \{a_0, a_1, a_2, a_3 \}^n$, with $a_p = p s$, with $p \in \{0, 1, 2, 3\}$ and where $s$ represents the distance between quantization levels and $p$ is the index of the quantized value.
We want to efficiently compute $w \cdot a$ using logical and bitwise operators, as in Equation~\ref{eq:binmult}. 
$a$ stores the activations of a neural model, so it is generally quantized to positive values (asymmetric quantization\footnote{\url{https://pytorch.org/blog/quantization-in-practice/}}).
This is due to the choice of the ReLU as an activation function, which zeros out negative inputs.
For the purpose of our technique, it is useful to re-map $a$ as a zero-centered symmetric distribution.
We scale $a$ by its mean. We obtain $\bar{a}$ as 
$$\bar{a_i} = a_i - \mu = a_i - \frac{b_0 + b_3}{2}  = a_i - \frac{3}{2} s$$ 
$\forall i = 0, \dots, n$. This converts the computation of $w \cdot a$ into
$$w \cdot a = \sum_{i=1}^{n}  w_i  (\bar{a}_i + \mu) = \sum_{i=1}^{n}  w_i  \bar{a}_i  + \sum_{i=1}^{n}  w_i  \mu.$$

The last term is the sum along the rows of the matrix weight scaled by $\mu$. Thus, it can be computed \emph{offline}, before the matrix multiplication starts, as both terms are known a priori. Let us consider the update function $c \leftarrow c \ + a  b$. We can initialize the matrix $c$ with these pre-computed row-wise terms rather than with zeros to avoid any impact of this computation on the performance.
Observe that $\bar{a_i} \in  \{\shortminus \frac{3s}{2} , \shortminus \frac{s}{2}, \frac{s}{2}, \frac{3s}{2}  \} $. 
Given that the multiplication by $\gamma$ and $s$ do not impact the performance, we need to multiply a $\{\shortminus 1, 1 \}$ binary vector against a 2-bit vector whose values are in $\mathcal{D}_a = \{\shortminus \frac{3}{2}, \shortminus \frac{1}{2}, \frac{1}{2}, \frac{3}{2}  \} $.
$\mathcal{D}_a$ is composed of two pairs of zero-symmetric values, namely 
$\{\shortminus \frac{3}{2}, \frac{3}{2}\}$ and $\{\shortminus \frac{1}{2}, \frac{1}{2}\}$. 
This suggests the possibility of employing a tailored representation made of multiple binary vectors. 
Hence, we represent $a$ using three binary vectors computed using the transformation
\begin{equation}
\label{app:eq:t_def}
\begin{gathered}
T : \{\mathcal{D}_a\}^n \rightarrow \{0, 1 \}^{n \times 3} \\ a \rightarrow \{t,h,m\}
\end{gathered}
\end{equation}

The definition of $T$ is reported in Table~\ref{app:tab:three_v_t}. Observe that, from now on, we will represent binary vectors using $\{0,1\}$ instead of $\{-1, 1\}$. This will ease the explanation of our approach, and it is also coherent on how binary vectors are actually represented during computation. The transformation $T$ generates two binary vectors $t,h$, and a binary mask $m$.
The names of the vectors suggest that $t$ ($t$hree halves) refers to  $\{\shortminus \frac{3}{2}, \frac{3}{2}\}$, while $h$ (one $h$alf) refers to $\{\shortminus \frac{1}{2}, \frac{1}{2}\}$.
\begin{table}[h]
    \centering
        \caption{Definition of the transformation $T$ mapping a 2-bits vector $a$ into its three binary vectors decomposition.}
    \label{app:tab:three_v_t}
    \begin{tabular}{lcccc}
        \toprule
        $a_i$ & $ \shortminus \frac{3}{2}$ & $\shortminus \frac{1}{2} $ & $\frac{1}{2}$ & $\frac{3}{2}$ \\
        \midrule
        $t_i$ & 0 & 0 & 0 & 1 \\
        $h_i$ & 0 & 0 & 1 & 0 \\
        $m_i$ & 1 & 0 & 0 & 1 \\
        \bottomrule
    \end{tabular}
    \vspace{0.1cm}

\end{table}

Two binary vectors, i.e., $t$ and $h$, are built according to the following rules.
\begin{itemize}
    \item a bit set to $1$ uniquely identifies the entry of the original vector $a$. This means that if $t_i=1 \Rightarrow a_i = \frac{3}{2}$ and if $h_i=1 \Rightarrow a_i = \frac{1}{2}$.

    \item a bit set to $0$ is intentionally ambiguous
    : $t_i =0 \Rightarrow a_i \in \{ \shortminus \frac{3}{2}, \shortminus \frac{1}{2}, \frac{1}{2}  \}$, $h_i =0 \Rightarrow a_i \in \{ \shortminus \frac{3}{2}, \shortminus \frac{1}{2}, \frac{3}{2}  \}$.
\end{itemize}
%
%
%
%
The mask $m$ is introduced to determine the uncertain entries of $t$ and $h$. We have that
\begin{itemize}
    \item $m_i = 1 \land t_i = 0 \Rightarrow a_i = \shortminus 3/2$;
    \item $m_i = 0 \land h_i = 0 \Rightarrow a_i = \shortminus 1/2$.
\end{itemize}
To ease the notation, we define the set of \emph{active entries} for $t$ as $E_t = \{ i\ | \ m_i = 1 \}$. The set of active entries for $h$ is $E_h = \{ i\ | \ \bar{m}_i = 1 \}$.
We highlight that $T$ is a bijective function that permits uniquely matching $a$ with its ternary representation $\{t, h, m \}$.
By using the mask $m$, we introduce a memory overhead of one bit for each value of $a$.
On the other hand, $m$ is necessary to split the $1/2$ multiplication into two $1/1$ multiplications. 
The idea is to multiply the binary vector $w$ with $t$ and $h$ respectively by involving in the computation only the active entries identified by the mask $m$. 
The considerations above inherently define a ternary operator $\texttt{mbm}(x,y,z)$, which we define on two generic binary vectors $x,y$, and a binary mask $z$.
In this case, $x$ plays the role of the weight vector, $y$ represents one between $t$ and $h$, while $z$ is the mask corresponding to $y$, i.e., $m$ and $\bar{m}$ when masking $t$ and $h$, respectively.
The set of active entries of $y$ is $E_y = \{ i\ | \ z_i = 1 \}$.

We now discuss how to implement the \texttt{mbm} operator.
First, let us focus on Equation~\ref{eq:binmult}.
It works by subtracting from the total number of bits involved in the computation, i.e., $n$, the number of times the bits of $u$ and $v$ present different values, multiplied by $2$.
Our approach to compute $\texttt{mbm}(x,y,z)$ is to multiply $x$ and $y$ using Equation~\ref{eq:binmult}, and then exploit the mask $z$ to fix the result.
There are two corrections to be applied.
The first one is on the number of bits actually involved in the computation that is identified by the cardinality of the active entries: $|E_y| =  \sum_i z_i$, which can be computed with a $\texttt{popcount}(z)$.
Now we need to count how many times $x$ and $y$ present different values in correspondence of active entries.
First, we compute $d = \texttt{xor}(x,y)$.  $d_i=1$ implies $x_i \neq y_i$.
These entries should be taken into account only if $z_i=1$, i.e., only in correspondence with active entries of the vector $y$.
This can be computed with \texttt{popcount}(\texttt{and}(\texttt{xor}($x$,\,$y$),\,$z$)).




Given these considerations, $\texttt{mbm}$ is defined as
\begin{multline}
\texttt{mbm}(x, y, z)  =  \texttt{popcount}(z) - \\ 
2 ~ \texttt{popcount}(\texttt{and}(\texttt{xor}(x,y),z)).\label{app:eq:mbm}
\end{multline}
The overall $1/2$ routine based on bitwise operations consists in applying the $\texttt{mbm}$ routine twice, first on the active entries of $t$ identified by $m$, and second on the active entries of $h$ identified by $\bar{m}$.
\begin{equation}
    \label{app:eq:1_2_matmul_dec}
    w \cdot a = \texttt{mbm}(w, t, m) + \texttt{mbm}(w, h, \bar{m}).
\end{equation}

As mentioned in Section~\ref{sec:bitwise}, \texttt{mbm} can be efficiently implemented using the \texttt{\_mm512\_ternarylogic\_epi64} instruction. This instruction allows to compute any ternary logic function, i.e., any logic function with three inputs, and has the same latency and throughput of the \texttt{xor} instruction.
That said, the cost of executing the \texttt{mbm} operator is the same as the one of computing Equation~\ref{eq:binmult}. 
$\texttt{popcount}(z)$ is the sum along the columns of a bit matrix. It runs in $\Theta(n^2)$ and its cost is negligible with respect to the cost of matrix multiplication, i.e., $\Theta(n^3)$. For this reason, we estimate that $tpp_{1/2}\simeq 2 \, tpp_{1/1}$.

\smallskip
\noindent \textbf{2/2}.
This quantization schema almost fills the effectiveness gap with full-precision networks~\cite{yang2020searching,zhou2016dorefa,gong2019differentiable,lee2021network}.
%
%
In this quantization schema, both the weight vector  and the activation vector 
are $2$-bit quantized vectors.
The naive $2/2$ MM can be obtained via an up-cast to $8/8$ that allows the use of the \fma instruction as discussed before.

We design an efficient alternative solution where both the weights and the activations are decomposed into three binary vectors.
The first step is to zero-center the two vectors by following the procedure described for the $1\text{/}2$ case. By doing that, we obtain the following $w$ and $a$. 
$$w = s_w\{ \shortminus \frac{3}{2}, \shortminus \frac{1}{2}, \frac{1}{2},  \frac{3}{2}  \}, \quad a = s_a\{ \shortminus \frac{3}{2}, \shortminus \frac{1}{2},  \frac{1}{2},  \frac{3}{2}  \} $$
where $s_w$ and $s_a$ represent the quantization step for $w$ and $a$, respectively.
The idea behind the $2\text{/}2$ routine is to apply the $1\text{/}2$ routine twice. 
Thus, the transformation $T$ is applied to both $w$ and $a$.
We obtain $ T(w) = \{t_w, h_w, m_w \} $ and $T(a) = \{t_a, h_a, m_a \}$.
The multiplication is decomposed as follows
\begin{itemize}
    \item the active entries of $t_w$, identified by $m_w$, are multiplied by the active entries of $t_a$, identified by $m_a$;
    \item the active entries of $t_w$, identified by $m_w$, are multiplied by the active entries of $h_a$, identified by $\bar{m}_a$;
    \item the active entries of $h_w$, identified by $\bar{m}_w$, are multiplied by the active entries of $t_a$, identified by $m_a$;
     \item the active entries of $h_w$, identified by $\bar{m}_w$, are multiplied by the active entries of $h_a$, identified by $\bar{m}_a$.  
\end{itemize}

Finally, the results are summed together.
We observe that for each one of the four bullets above, we have two different sets of active entries.
The actual set of active entries is given by their intersection.
This is implemented by a logical \texttt{and} between the two masks involved.
To summarize, our novel routine for $2\text{/}2$ MM allows to compute $w \cdot a $ as
\begin{align*}
\label{app:eq:2_2mult}
w \cdot a  = 
& ~ \texttt{mbm}(t_w, h_w, m_w \land m_a) ~ + \\
& ~ \texttt{mbm}(t_w, h_a, m_w \land \bar{m}_a) ~ + \\ 
& ~ \texttt{mbm}(h_w, t_a, \bar{m}_w \land m_a) ~ + \\
& ~ \texttt{mbm}(h_w, h_a, \bar{m}_w \land \bar{m}_a).
\end{align*}
Hence, $tpp_{2/2}\simeq 4 \, tpp_{1/2} \simeq 8 \, tpp_{1/1}$.

\section{Experimental Evaluation}
\label{sec:exp}  

In this section, we present our experimental evaluation of \apb. Our empirical analysis consists of two different steps. In the former one (Section~\ref{subsec:comprperf}), we assess the memory compression capabilities of \apb by comparing it against i) pruning, ii) quantization, and iii) pruning combined with quantization/binarization for convolutional neural networks. In the latter subsection (Section~\ref{subsec:effeval}), we perform an efficiency analysis by evaluating the performance of our novel low-bits matrix multiplication algorithms against existing solutions on CPU. Moreover, we leverage our routines to compare the execution time of convolutional neural networks at different bit-width and compare them to \apb-compressed models.

\subsection{Experimental Setup}
\label{subsec:expsetup}

\smallskip
\noindent \textbf{Datasets.}
We comprehensively evaluate \apb against several state-of-the-art competitors on two widely adopted datasets for Image Classification, namely \cifar~\cite{cifar10} and \imagenet~\cite{imagenet_cvpr09}. The \cifar dataset consists of a set of $60$K $32\times32$ images split into $50$K and $10$K training/test samples, respectively, labeled with $10$ classes. The \imagenet dataset consists of $1.2$M training images and about $50$K test images labeled with $1\text{,}000$ classes. 

\smallskip
\noindent \textbf{Network Architectures}.
We evaluate the performance of \apb in compressing different kinds of convolutional networks. In particular, these architectures are ResNet-18/20/56~\cite{DBLP:journals/corr/HeZRS15} and VGG-Small~\cite{simonyan2014very} on the \cifar dataset, and ResNet-18/34/50 and WideResNet-50 on \imagenet, which are the benchmark architectures for quantization methods. We apply \apb on all the layers of the networks, except the first, the last, and the downsampling layers, unless differently specified. 
The effectiveness of the models is measured in terms of Top1 classification accuracy.

\smallskip
\noindent \textbf{Training Details}.
We apply \apb after initializing the network weights with pre-trained models. For each layer $i$, we set $\alpha_i = \mu_i$ and $\delta_i = 3\sigma_i$, where 
$\mu_i$ is the mean of $|w_i|$ and $\sigma_i$ is the standard deviation of $w_i$.
Under the assumption that $w_i \sim \mathcal{N}(\mu,\,\sigma^{2})$~\cite{DBLP:journals/corr/HanPTD15}, this ensure high compression rate at initialization.
We employ an SGD optimizer with a cosine annealing learning rate scheduler. On \cifar, we train for $500$ epochs, with a batch size of $128$ and a learning rate of $1\mathrm{e}{\shortminus2}$.
On \imagenet, we train for $100$ epochs when using real-values activations and $150$ for $2$-bits activations, with batch size $256$ and learning rate $1\mathrm{e}{\shortminus3}$.
We freeze the value of $\alpha$ and $\delta$ when reaching half of the epochs of each training to allow fine-tuning of the surviving values. \apb is implemented in PyTorch~\cite{NEURIPS2019_9015}.

\smallskip
\noindent \textbf{Weight Decay}. The compression aggressiveness of \apb can be controlled with the weight decay $\lambda$, namely the $L_2$ penalty applied on the weights of a neural network at training time. 
The larger the weight decay, the lower (on average) the absolute value of the trained parameters. As we do not apply the weight decay to $\alpha$ and $\delta$, the amplitude of the binarization interval is not directly affected by its value. As a consequence, increasing $\lambda$ will force more weights to fall into the binarization interval, thus incrementing the percentage of binary values w.r.t to full-precision ones and delivering higher compression.

\begin{table}[t]
\centering
\caption{Comparison between \apb and state-of-the-art CL competitors in terms of weights bit width and Top1 accuracy. Note that CL methods do not compress the first, the last, and downsample layers. To ease the comparison with Table~\ref{tab:memefficiency_alc}, we report both the CL and AL weights bit-width using the format CL (AL). 
\label{tab:memefficiency}}
\adjustbox{max width=\columnwidth}{
\begin{tabular}{lllrr}  
 \toprule

        Dataset & Network & Method & \thead{Weights \\ Bit width \\ CL (AL)}& \thead{Top1 \\ ($\%$)}\\
        \midrule
        
        \multirow{12}{*}{\cifar} & \multirow{5}{*}{ResNet-20} & FP &32.0 & 93.6 \\
        \cmidrule{3-5}
        & & IR-Net&  1.0 (1.4) & 90.8\\
        & & SLB&  2.0 (2.4) & 92.0  \\
        & & \apb (Ours) & 1.0 (1.4) & 92.3 \\
        \cmidrule{2-5}
        & \multirow{5}{*}{VGG-Small} & FP & 32.0 & 94.5 \\
        \cmidrule{3-5}
        & & SLB & 1.0 (1.6)& 93.8 \\
        & & SLB & 2.0 (2.6)& 94.0 \\
        & & \apb (Ours) &   1.0 (1.6)& 94.3 \\
        \cmidrule{2-5}
        &  \multirow{3}{*}{ResNet-18} & FP &32.0 & 95.4 \\
        \cmidrule{3-5}
        & & MPT & 1.0 (1.5)& 94.8 \\
        & & \apb (Ours) & 1.0 (1.5)& 95.0 \\
        \midrule
        
        \multirow{15}{*}{\imagenet} & \multirow{5}{*}{ResNet-18} & FP &32.0 & 69.6 \\
        \cmidrule{3-5}
        & & EWGS & 1.0 (3.0)& 67.3\\
        & & EWGS & 2.0 (4.0)& 69.3\\
        & & \apb (Ours) & 1.4 (3.4)& 69.2 \\
        & & \apb (Ours) & 1.7 (3.7)& 69.7 \\
        \cmidrule{2-5}
        &\multirow{4}{*}{ResNet-34} & FP & 32.0 & 73.3 \\
         \cmidrule{3-5}
        & & EWGS & 1.0 (2.1)& 72.2 \\
        & &  \apb (Ours)  & 1.4 (2.5) & 73.2 \\
        \cmidrule{2-5}
        &\multirow{3}{*}{ResNet-50} & FP & 32.0 & 76.1 \\
         \cmidrule{3-5}
        & &  \apb (Ours)  & 1.1 (7.3) & 75.8 \\
       \cmidrule{2-5}

         & \multirow{4}{*}{WideResNet-50} & FP & 32.0 & 78.5 \\
          \cmidrule{3-5}
          & & MPT & 1.0 (3.3) & 74.0 \\
        & &  \apb (Ours)  & 1.1 (3.4)  & 77.6 \\
        
\bottomrule
\end{tabular}
}
\end{table}

\subsection{Compression Performance of \apb with 32-bits activations}
\label{subsec:comprperf}
In this section, we evaluate the performance of \apb as a memory compression framework. For this purpose, we consider models whose activations are kept in $32$-bits. Recall that, even when weights are binarized, maintaining activations in full precision prevents any inference advantage, as discussed in Section~\ref{sec:bitwise}, paragraph ``$1 \text{/} 32$''.

We compare \apb against several different techniques, namely i) quantization, ii) pruning, iii) combination of pruning and binarization.
Compression methods adopt two main approaches. The first one, which is adopted, for example, by pruning techniques, compresses all the network layers.
We call these methods \emph{All-Layers Compressors} (AL). The second one, which is adopted by binarization, low-bits quantization, and pruning + binarization, works by leaving in full precision the first, the last, and the downsample layers, if any. We name these solutions \emph{Convolutional Layers Compressors} (CL).  

As already mentioned in Section \ref{subsec:expsetup}, \apb belongs to CL methods. Despite that, in our experiments, we compare it against methods belonging to both families. To ease the comparison, we report the results in two different tables. Table~\ref{tab:memefficiency} reports the results for AL methods, while Table~\ref{tab:memefficiency_alc} reports the results for CL methods. In Table~\ref{tab:memefficiency}, we report the CL weights bit-width and the AL weight bit-width to allow a direct comparison between the two tables. 
Both for AL and CL, we measure the memory impact as average \emph{weights bit width}, namely the ratio between the number of bits required to store a model and its parameters.
When comparing against CL methods, the first, the last, and the downsample layers of the network are excluded from the computation, while these are included when comparing against AL methods.

\subsubsection{Compression of Convolutional Layers (CL)}
We start our comparison with methods compressing only the Convolutional Layers (CL).

\begin{table}[t]
\centering
\caption{Comparison between \apb and pruning in terms of compression rate width and Top1 accuracy.  Here, we list All Layers Compressors (AL) methods, which compress the first, the last, and downsample layers. BB~\cite{van2020bayesian} uses mixed-precision activations. \apb marked with \ddag\xspace indicates the usage of 2-bit activations, while \dag\xspace  indicates that the downsample layers are compressed as well. \label{tab:memefficiency_alc}} 
\adjustbox{max width=\columnwidth}{
\begin{tabular}{lllrr}  
        \toprule
        Dataset & Network & Method & \thead{Weights \\ Bit width \\ AL}& \thead{Top1\\ ($\%$)}\\
        \midrule
        
        \multirow{7}{*}{\cifar} & \multirow{3}{*}{ResNet-20}  & FP &32.0 & 93.6 \\
        \cmidrule{3-5}

        & & Pruning~\cite{renda2019comparing} &4.8 & 91.1 \\
        & & $\mathtt{APB}$ (Ours) & 1.4 & 92.3 \\
        \cmidrule{2-5}
        & \multirow{4}{*}{ResNet-56} & FP &32.0 & 94.6 \\
        \cmidrule{3-5}
        & & Pruning~\cite{renda2019comparing} & 4.7 & 93.9 \\
        & & Pruning~\cite{renda2019comparing} & 1.0 & 91.9 \\
        & & $\mathtt{APB}$ (Ours) & 1.3 & 93.6  \\
        \midrule
        \multirow{12}{*}{\imagenet} & \multirow{7}{*}{ResNet-18} & FP & 32.0 & 73.3  \\
        \cmidrule{3-5}

        & &  BB  & 5.1  & 69.5 \\
        & & BB & 4.1  & 68.1 \\ 

        & &  \apb (Ours) & 3.5 & 69.7 \\
        & &  \apb + PTQ (Ours) & 2.4 & 69.6 \\
        & &  \apb\textsuperscript{\!\ddag}+ PTQ (Ours) & 2.8 & 67.4 \\

        \cmidrule{2-5}
        &  \multirow{6}{*}{ResNet-50} & FP &  32.0 & 76.1 \\
        \cmidrule{3-5}

        & & Pruning~\cite{renda2019comparing} & 5.3 & 75.8 \\
        & & \apb
         (Ours) & 7.3 &  75.6\\
         & &  \apb + PTQ (Ours) & 5.4 & 75.6 \\
         & &  \apb\textsuperscript{\!\dag}+ PTQ (Ours) & 2.1 & 75.3 \\

\bottomrule
\end{tabular}
}

\end{table}

\smallskip
\noindent \textbf{Comparison with quantization approaches.}
We compare \apb against state-of-the-art neural quantization techniques on \cifar and \imagenet
and report the results in Table~\ref{tab:memefficiency}. All the low-bit quantization techniques belong to the CL category described above. 
For each quantization scheme, we report the best-performing competitor. The state-of-the-art on \cifar is SLB~\cite{yang2020searching} for all quantization schemes except the $1\text{/}32$ case on ResNet-20, where the best performance is achieved by IR-Net~\cite{qin2020forward}.
On \imagenet, the state-of-the-art solution is
EWGS~\cite{lee2021network} for both ResNet-18 and ResNet-34.

Results show that \apb outperforms all other state-of-the-art quantization approaches by providing a superior solution in terms of space and accuracy.
Indeed, \apb achieves higher accuracy than the $2$-bits quantization while saving up to $2\times$ memory on \cifar and $1.4\times$ on \imagenet.
In comparison to $1$-bit quantization, \apb presents a negligible memory overhead on \cifar.
In fact, the surviving full-precision values for these models account for about $0.1\%$ of the total weights.
On \imagenet, the number of surviving full-precision weights required by \apb is larger, as demonstrated by the $0.4$ more bits per weight.
However, this memory overhead is strongly counterbalanced by $1.9$ and $1.0$ Top1 accuracy points of gain for ResNet-18 and ResNet-34, respectively, compared to $1$-bit quantization. On ResNet-18, our models even outperform $2$-bits quantization by $0.4$ Top1 accuracy points despite its reduced memory footprint. 

\smallskip
\noindent \textbf{Comparison with combinations of pruning and binarization.} Recently, some works explore the combination of pruning \& binarization~\cite{diffenderfer2020multi,li2022equal}. These methods adopt an orthogonal approach with respect to \apb. While with \apb weights are either binary or full-precision, with existing approaches combining pruning and binarization, the weights are either binary or \emph{zero}. We compare the performance of \apb with respect to Multi Prize Ticket (MPT) \cite{diffenderfer2020multi} in terms of memory compression. MPT works by discovering highly effective sparse sub-networks in sufficiently over-parameterized neural networks without the need for further training. Moreover, they apply the \emph{sign} function to binarize the surviving weights, thus providing a sparse and binary network.

We argue that sparsification does not provide memory advantages if the non-zero weights are binary. Consider a tensor of size $n$ with $nnz$ non-zero entries. The cost of storing a non-zero entry is given by the number of bits to its value $b_v$ plus the number of bits to save its position $b_p$. 
In this case, i.e., $b_v = 1$, as surviving weights are binarized. In MPT~\cite{diffenderfer2020multi},
the authors propose a sparsification ratio of $80\%$ for this architecture. This means that one weight out of five is stored. Storing $5$ weights in pure binary format only costs $5$ bits, so
if $b_p > 4$, the sparse-binary format does not provide memory footprint advantages compared to the pure binary one. Due to the large size of network layers, $4$ bits are not enough to store the indexes of non-zero values.
A more memory-efficient approach would be to store one bit for every weight to indicate whether the corresponding entry is pruned. In this case, the total memory impact would be $n+nnz$, which we approximate with $n$ for simplicity.

Table~\ref{tab:memefficiency} reports the evaluation results of \apb against MPT. \apb outperforms the performance of MPT both on ResNet-18 for \cifar, and for WideResNet-50 on \imagenet. 
For ResNet-18 in \cifar, \apb matches the memory compression of MPT and also delivers a slight ($0.2$) effectiveness improvement.
For WideResNet-50 on \imagenet, the model learned with \apb achieves more than $3$ points of Top1 Accuracy improvement w.r.t. MPT. with a memory overhead of only $0.1$ bits per weight.

\subsubsection{Compression of All Layers (AL)}
We now compare \apb to methods compressing the whole network. These approaches also tackle the first, the last, and the downsample convolutional layers. Batch Normalization layers are left in full precision as their impact is negligible, as they account for less than $0.5\%$ of network parameters in all the evaluated architectures.

Table~\ref{tab:memefficiency} shows that the impact of the layers left out by CL approaches can be considerable. On \imagenet, we observe that the non-compressed parameters account for an overhead of $2$ bits per weight for ResNet-18 and $6.2$ for ResNet-50.
These numbers are obtained as the difference between the CL and the AL value in the ``Weights Bit width CL (AL)'' column of Table~\ref{tab:memefficiency}.
On ResNet-18, we observe that the $70\%$ of the overhead depends on the last fully connected layer. We experimentally verify that these last fully-connected layers can be quantized to 8-bit integers with a simple post-training quantization (PTQ)
solution, that converts the weights from a floating point 32 representation to 8-bit integers, \footnote{\url{https://pytorch.org/docs/stable/quantization.html}}
without harming the Top1 accuracy of the model. This halves the overhead of the non-compressed layers, reducing it to $0.9$ bits per weight. For ResNet-50, quantizing the last fully connected layer to $8$-bits allows reducing the overhead to $4.3$ bit per weight. 




\smallskip
\noindent \textbf{Comparison with pruning.}
We compare the memory compression capabilities of \apb against state-of-the-art pruning techniques, namely magnitude-based pruning with learning-rate rewinding~\cite{renda2019comparing}. We report the results of this comparison in Table~\ref{tab:memefficiency_alc}.
In the original article~\cite{renda2019comparing}, the authors express the compression rate as the inverse of the sparsity ratio, i.e., $5\%$ sparsity corresponds to $20\times$ compression rate. Indeed, this does not account for the space required to store the indexes of the nonzero entries. 
We recompute the weights' bit width by taking it into account. 
We compare \apb against pruning on ResNet-20, ResNet-56 on \cifar, and on ResNet-50 on \imagenet. On \cifar, \apb vastly outperforms pruning in terms of memory/accuracy trade-off. When compressing ResNet-20, \apb delivers a model which is $3.4\times$ smaller but $1.2$ points more accurate. 
Regarding ResNet-56, \apb can deliver the same level of accuracy of a sparsified model with a $3.3\times$ memory saving or the same level of memory compression with $1.7$ points of Top1 accuracy improvement.
On \imagenet, we observe that \apb combined with Post Training Quantization (PTQ)
on the last layer allows matching the performance of pruning using ResNet-50 as the backbone. Furthermore, we perform an experiment where we apply \apb on the downsample convolutional layer of this architecture, marked with \dag\xspace  in Table~\ref{tab:memefficiency_alc}. For this model, the downsample convolutional layers account for the $10\%$ of the total memory impact. Thus, we can obtain a model that only suffers from $0.5$ accuracy degradation compared to pruning but allows saving up to $2.5\times$ memory footprint.

\smallskip
\noindent \textbf{Comparison with combinations of pruning and quantization.}
We compare with Bayesian Bits (BB)~\cite{van2020bayesian}, a state-of-the-art compression approach mixing quantization at different bit widths with pruning. In detail, we compare in terms of bit width of their methods against our models learned with \apb. As for the other experiments, we use ResNet-18 on Imagenet as the model for comparison and report the results in Table~\ref{tab:memefficiency_alc}.
As mentioned, we also apply PTQ on the final fully connected layer of ResNet18. Compared to BB, \apb can deliver the same accuracy but allows us to save up $2.1\times$. For the sake of fairness, we point out that activations in BB are quantized to mixed precision ($2/4$ bits), which can decrease the model's accuracy. We also compare to a model compressed with \apb whose activations are quantized to $2$-bits, marked with the \ddag\xspace symbol in Table~\ref{tab:memefficiency_alc}. Even if employing reduced bit width for activations, this model gains a $1.5\times$ memory savings with reduced performance degradation.

\subsection{Efficiency Evaluation}
\label{subsec:effeval}
In this Section, we provide an extensive efficiency evaluation of our low-bits matrix multiplication routines and of 
\apb-compressed networks. First, we assess the performance of our low-bits matrix multiplication routines, comparing them against state-of-the-art dense matrix multiplication frameworks. Then, we show that our \apb compressed networks outperform existing quantization methods in terms of efficiency-accuracy trade-off. 

\begin{figure}[t]
\centering
\includegraphics[width=\columnwidth]{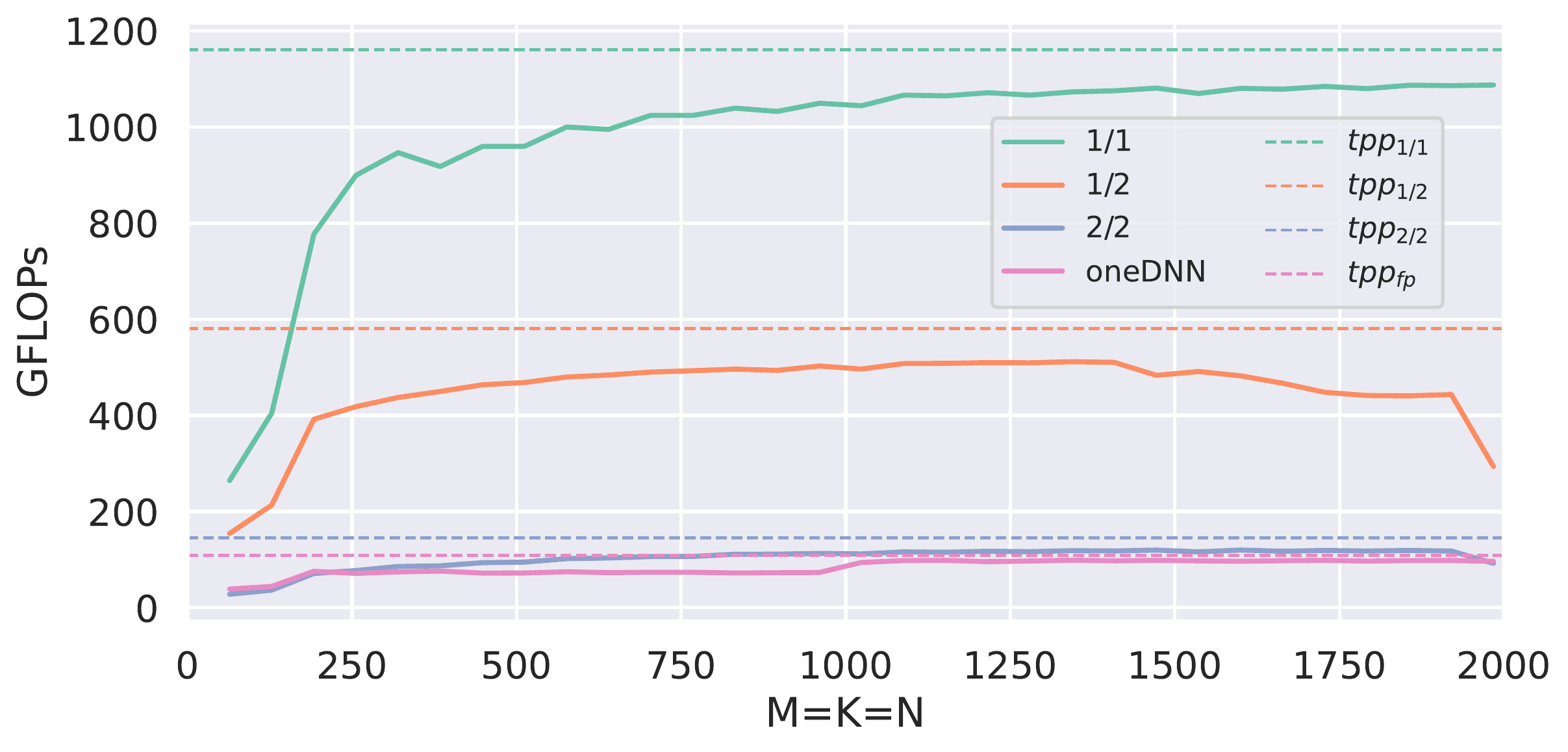}
\caption{GFLOPs performance analysis of our novel bitwise MM routines against the oneDNN library used for dense MM. The analysis is performed on square matrices, i.e., $M$ = $K$ = $N$, where $M$ is the number of rows of the first operand, $K$ is the shared dimension, and $N$ is the number of columns of the second operand.}
\label{fig:gflops}
\end{figure}

\smallskip
\noindent \textbf{Low-bits Matrix Multiplication.}
We now compare the efficiency of our Matrix Multiplication (MM) routines, i.e., 1/1, 1/2, 2/2, against state-of-the-art highly-optimized MM CPU libraries. In detail, we measure the achieved GFLOPs at different sizes of the matrices and we compare them to their $tpp$. $tpp$ is computed according to Equation~\ref{eq:tpp} and it is reported in Figure~\ref{fig:gflops} as dotted lines. Experiments are conducted on an Intel Xeon Gold 5318Y CPU, clocked at $3.4$ GHz and equipped with the \textsc{avx-512} instruction set.
Our novel inference routines are written in C++ and compiled with the -O3 option using GCC 11.2.0, with single-threaded execution. 
Figure~\ref{fig:gflops} reports an experimental comparison of square matrices. We report the size of the matrices under multiplication, i.e.,  on the $x$-axis. We include in the comparison the performance of dense matrix multiplication as implemented in the oneDNN library. This library is the state-of-the-art solution for dense multiplication, employing industrial-level optimizations such as Just In Time (JIT) code compilation. 
Figure~\ref{fig:gflops} shows how large-enough matrices allow all the tested algorithms, i.e., our bitwise MM routines and the oneDNN library for dense MM, to achieve their best performance and get close to their $tpp$.
The experimental results show that the $1\text{/}1$, $1\text{/}2$, and $2\text{/}2$ routines reach up to $93\%$, $88\%$, and $85\%$ of their theoretical peak performance, respectively. For large enough matrices, oneDNN GFLOPs range between the $86\%$ and the $90\%$ of the $tpp$ of dense MM.
The results reported in Figure~\ref{fig:gflops} confirm that our bitwise multiplication routines are properly implemented, given that the gap with $tpp$ is within the $15\%$. Further optimizations may be employed, such as blocking strategies or micro-kernel parameters optimization according to the CPU architecture. These are optimization strategies that are commonly applied in the dense case~\cite{goto2008anatomy}. Although interesting, these optimizations go beyond the scope of this work, and we leave their investigation as future work.

Square matrices allow MM routines to get close to their $tpp$ at any bit width. Indeed, in DNN inference, rectangular matrices are much more common than squared ones. To evaluate the efficiency of our novel matrix multiplication algorithms on a real use case, we test them on the shapes obtained by applying the \emph{im2col} transformation on the layers of ResNet-like architectures.
We include the $8\text{/}8$ MM  in our analysis by employing the implementation provided by the FBGEMM~\cite{khudia2021fbgemm} library.

\begin{figure}[t]
\centering
\includegraphics[width=\columnwidth]{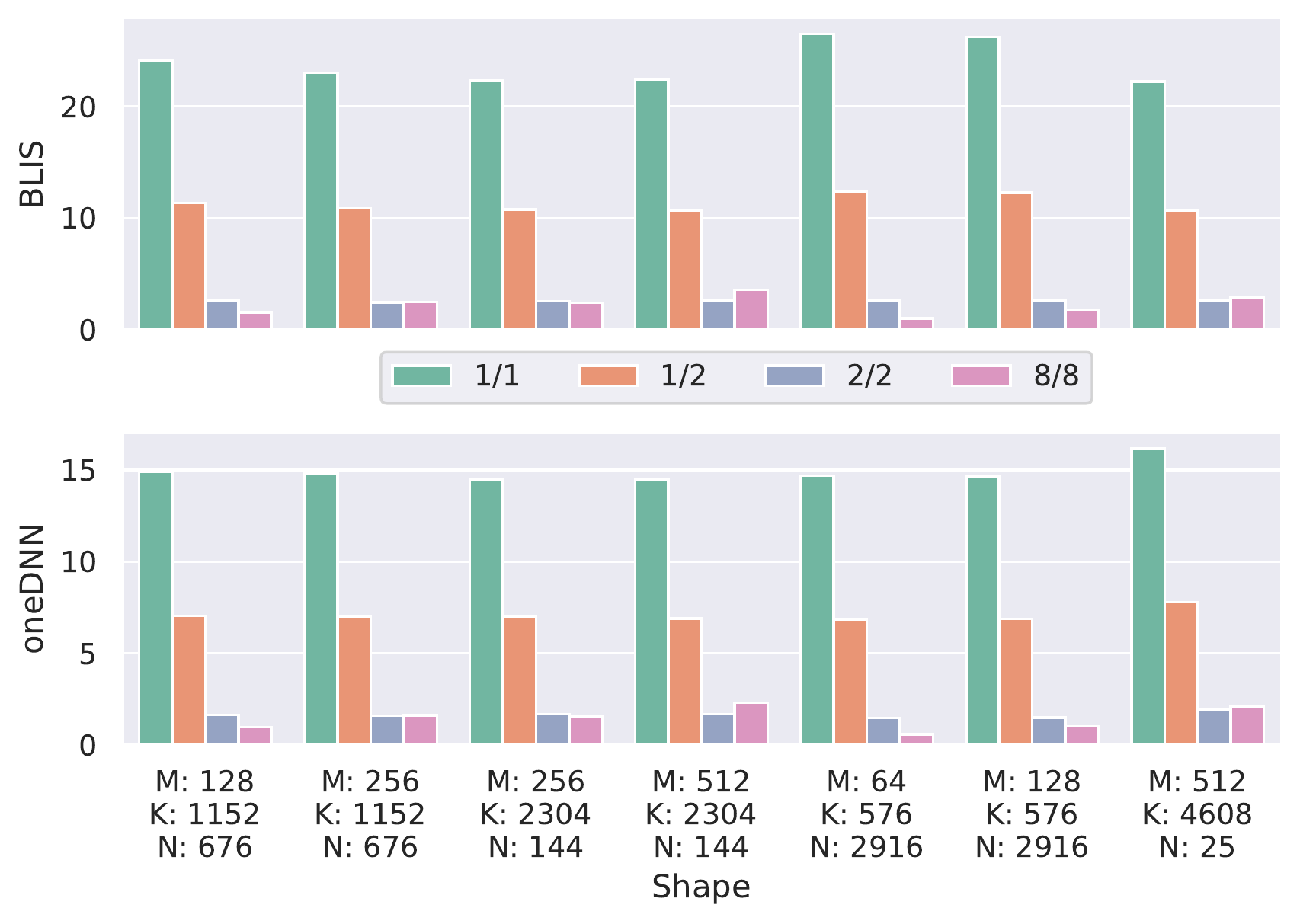}
\caption{Comparison between matrix multiplication routines at different quantization levels. $M$ is the number of rows of the first operand, $K$ is the shared dimension, and $N$ is the number of columns of the second operand.}
\label{fig:speeup_blis_dnnl}

\end{figure}

Results are reported in Figure~\ref{fig:speeup_blis_dnnl}, where the $x$-axis indicates the shapes of the matrices under evaluation and the $y$-axis reports the speedup compared to the dense MM. We adopt two different implementations for dense MM. The first one is OneDNN, as in Figure~ \ref{fig:gflops}. BLIS~\cite{huang2016blislab} is an open-source GEMM library with assembly-level architecture-tailored optimizations and presents an optimization level closer to our C++ implementation. Observe that, on the shapes under evaluation, oneDNN is 60\% faster than BLIS, which is a remarkable speedup considering that they employ the same algorithm for matrix multiplication.

Figure~\ref{fig:speeup_blis_dnnl} shows that our novel MM routines are significantly faster than dense multiplication. For example, the $1\text{/}1$ routine delivers up to $15\times$ speedup compared to oneDNN and up to $25\times$ compared to BLIS. 
In this case, we witness a speedup larger than the theoretical estimated one, i.e., $10.5\times$.
This is caused by the poor performance of dense MM on rectangular-shaped matrices~\cite{nardini2022distilled}. In fact, rectangular matrices do not allow fully masking the memory-bounded nature of the MM operation.\footnote{\url{https://en.algorithmica.org/hpc/algorithms/matmul/}}
This is evident in Figure~\ref{fig:speeup_blis_dnnl}, where, 
on the shapes under evaluation, oneDNN reaches between the $50\%$ and the $70\%$ of $tpp$. Conversely, $1\text{/}1$, $1\text{/}2$ and $2\text{/}2$ deliver respectively at least the $85\%$, $80\%$ and $75\%$ of their theoretical peak performance.

Overall, the forward pass on ResNet-18 achieves $6.85\times$ and $1.5\times$ speedup compared to FBGEMM, the state-of-the-art MM library for quantized models on CPU, when employing the $1\text{/}2$ and $2\text{/}2$ MM routines, respectively. To the best of our knowledge, FBGEMM is the best available solution for MM at low bit widths, excluding our novel routines. 
Our solution allows, for the first time, to efficiently exploit the plethora of different quantization methods for $1\text{/}2$ and $2\text{/}2$ directly on CPU. Also, the involved algorithms are novel and can be implemented on all the available computing platforms, such as GPUs or FPGAs. 
We also observe that the performance ratio between different low-bits implementation precisely matches the predictions we made in Section~\ref{sec:bitwise}: the $1\text{/}2$ quantization schema is $2\times$ slower than the $1\text{/}1$, while the $2\text{/}2$ is $8\times$ slower.

\noindent \textbf{Efficiency/Accuracy Trade-offs}.
We now compare the efficiency-accuracy trade-off that different low-bits quantization methods deliver compared to \apb.
We perform the analysis using the ResNet-18 architecture on the \imagenet dataset.
We compute the inference time as the total MM time.
The time spent on the first and the last layer, as well as on the batch-normalization and down-sampling layers, is ignored as 
these layers are not compressed in any of the approaches --- as in Table~\ref{tab:memefficiency}.
The comparison involves networks quantized with the $1$/$1$, $1$/$2$, and $2$/$2$ configurations.
The $1\text{/}1$ accuracy is achieved by using SA-BNN~\cite{liu2021sa}, while $1\text{/}2$ and $2\text{/}2$ are achieved by using EWGS~\cite{lee2021network}. Both methods are state-of-the-art low-bits quantization approaches. We do not include methods such as ReActNet~\cite{liu2020reactnet} or ElasticLink~\cite{hu2022elastic} that leverage custom architectures. We observe that \apb could be easily employed in conjunction with these approaches, which is also part of our future work.
In this analysis, \apb is used to compress the weights of the network, while activations are quantized to $2$ bits using the EWGS quantizer.
We also experiment \apb with $1$-bit quantized activations, and we achieved worse performance than $1\text{/}2$ quantization. This aspect is detailed in the subsequent paragraph.

The inference time reported for \apb is the sum of the time for $1\text{/}2$ MM and the time for sparse-dense MM, as shown in Equation~\ref{eq:mmdistprop}. LIBXSMM~\cite{heinecke2016libxsmm} is used for sparse-dense matrix multiplication. 

\begin{figure}[t]
\centering
\includegraphics[width=\columnwidth]{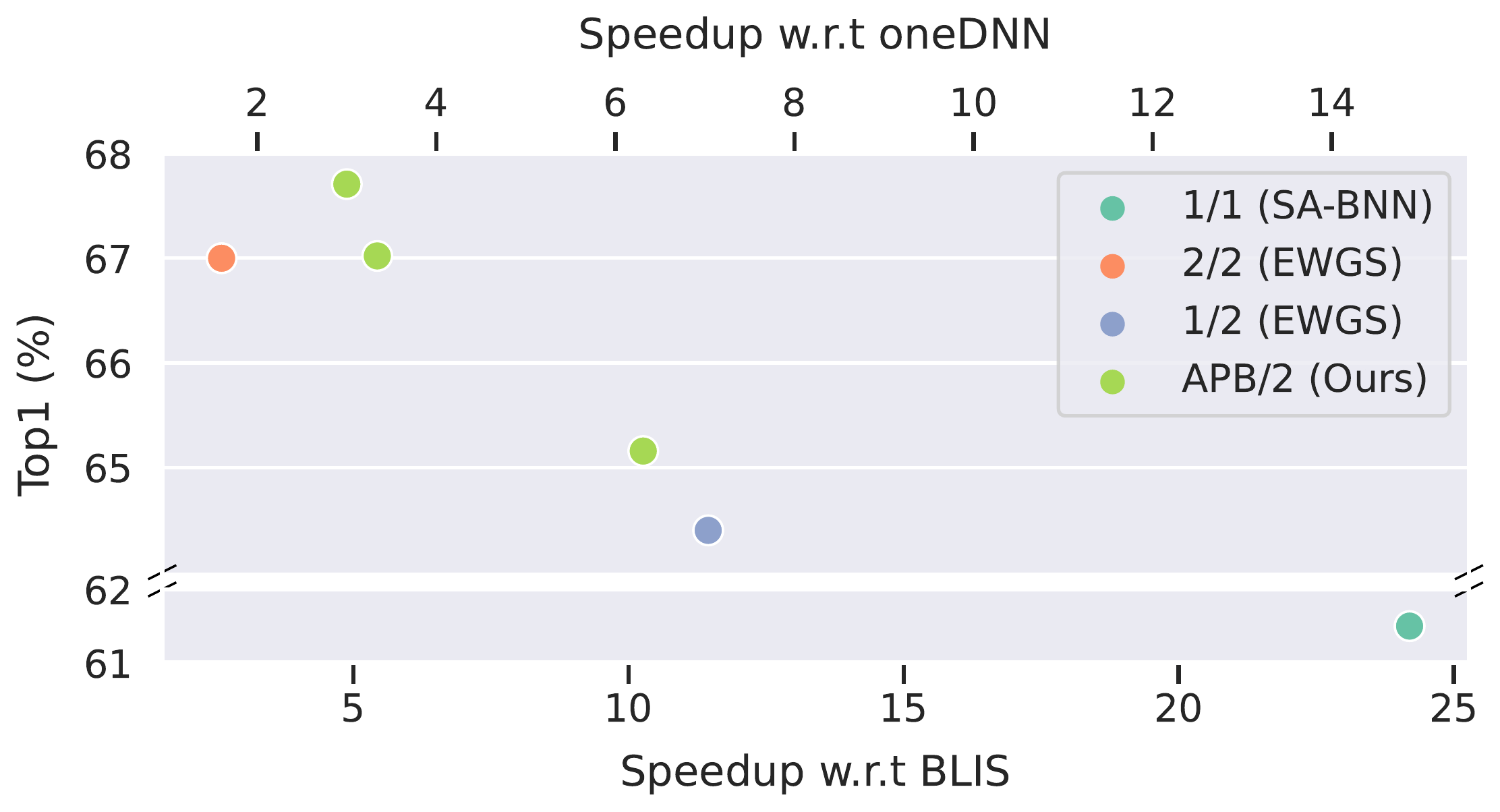}
\caption{Efficiency/Accuracy trade-offs of ResNet-18 on \imagenet using different quantization schema.}
\label{fig:efficiency}

\end{figure}

Figure~\ref{fig:efficiency} reports the results of the comparison in terms of speedup w.r.t. both BLIS and oneDNN ($x$-axis) and Top1 accuracy ($y$-axis). 
\apb largely dominates over $2\text{/}2$ quantization provided by EWGS.
Our models can deliver $1.9\times$ speedup with $0.7$ Top1 accuracy improvement or $2.0\times$ speedup at the same accuracy level of the $2\text{/}2$ quantized model.
The sparsity of surviving full-precision values per layer spans in $96\text{-}99 \%$. In this range, we experimentally verify that sparse MM is always faster than the $2/2$ routine, it matches the performance of $1/2$ at about $97\%$, and it is faster than $1/1$ at $99\%$.
Moreover, thanks to our novel MM routine, the $1\text{/}2$ models offer more than $11\times$ speedup with a performance degradation of $8\%$ w.r.t. to the full-precision model. 
Regarding $1\text{/}1$, its extreme speedup asks for a significant accuracy degradation, i.e., $7$ points of Top1 accuracy compared to the original full-precision model.

\smallskip 
\noindent \textbf{Comparison with pure binarization.}
We experiment with the combination of \apb with $1$-bit activations. We observe that the performance of our models outperforms by margin pure binary networks, by reaching $63.0$ of Top1 accuracy. Anyway, this effectiveness improvement comes at the price of an elevated number of surviving full-precision weights, reaching $8\%$ in some cases. 
The cost of these full-precision weights at inference time matches or sometimes surpasses the cost of binary multiplication.
In practice, this means that the $1\text{/}2$ scenario offers superior performance in terms of efficiency/accuracy tradeoff. This is coherent with the network quantization literature, where different works prove that reducing the precision of the activation worsens the effectiveness of the models more than reducing the precision of the weights

\section{Conclusions and Future Work}
\label{sec:concl}
We proposed \apb, a novel compression technique that merges binarization and pruning together to exploit the benefits provided by these two orthogonal techniques. We showed that \apb jointly maximizes the accuracy achieved by the network while minimizing its memory impact by identifying an optimal partition of the network parameters among these two sets. Furthermore, we presented two novel matrix multiplication algorithms for extreme low-bit configurations, namely $1\text{/}2$ and $2\text{/}2$, where $1 \text{/} 2$ refers to binary weights and 2-bit activations, while in the $2 \text{/} 2$ configuration both weights and activations are quantized to $2$ bits.
We performed a comprehensive experimental evaluation on two widely-adopted benchmark datasets, i.e., \cifar and \imagenet. 
Experiments show that \apb achieves better accuracy/memory trade-off w.r.t. to state-of-the-art compression methods based on i) quantization, ii) pruning, and, iii) the combination of pruning and quantization. 
Our novel matrix multiplication routines deliver a major speedup compared to the existing solution for low-bits matrix multiplication on CPU, ranging from $6.9\times$ for the $1\text{/}2$ configuration to $1.5\times$ for the $2\text{/}2$ configuration.
Moreover, the experimental results show that \apb is $2\times$ faster than the 2-bit quantized model with no loss in accuracy.
On the one hand, our novel matrix algorithms open up to exploiting quantized networks on CPU. Also, they may boost the investigation of $1\text{/}2$ quantization scenario, given that a very fast inference engine is available. On the other hand, we show that \apb-compressed networks, where binary and full-precision weights are mixed in the same weight tensor, allow for better performance compared to fixed quantization, e.g., $2$-bits. The importance of a reduced portion of full-precision weights, evidenced by pruning techniques in previous work, is stressed again in this new hybrid format.

As future work, we plan to extend \apb to automatically identify the optimal quantization schema for each layer activation to improve the efficiency/accuracy trade-off. 
This would be possible due to the availability of efficient matrix multiplication routines covering several quantization schemas. Moreover, we are interested in applying \apb in conjunction with highly effective custom architectures, such as ElasticLink~\cite{hu2022elastic}. We also plan to extend the evaluation of \apb to transformer-based neural architectures.

\section*{Acknowledgments}
Funding for this research has been provided by: MUR-PRIN 2022 ``Algorithmic Problems and Machine Learning''; PNRR - M4C2 - Investimento 1.3, Partenariato Esteso PE00000013 - ``FAIR - Future Artificial Intelligence Research'' - Spoke 1 ``Human-centered AI'' funded by the European Union (EU) under the NextGeneration EU programme; the EU's Horizon Europe research and innovation programme EFRA (Grant Agreement Number 101093026).  Views and opinions expressed are however those of the author(s) only and do not necessarily reflect those of the EU or European Commission-EU. Neither the EU nor the granting authority can be held responsible for them.

\bibliographystyle{IEEEtran}
\bibliography{paper}

\begin{IEEEbiography}[{\includegraphics[width=1in,height=1.25in,clip,keepaspectratio]{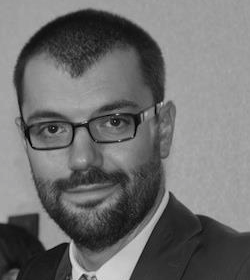}}]{Franco Maria Nardini}
is a senior researcher with the National Research Council of Italy. His research interests focus on machine learning and Web information retrieval. He authored more than 80 papers in peer-reviewed international journals and conferences. He received the Best Paper Award at ACM SIGIR 2015.
For more information: \href{http://hpc.isti.cnr.it/~nardini}{http://hpc.isti.cnr.it/~nardini}.
\end{IEEEbiography}

\begin{IEEEbiography}[{\includegraphics[width=1in,height=1.25in,clip,keepaspectratio]{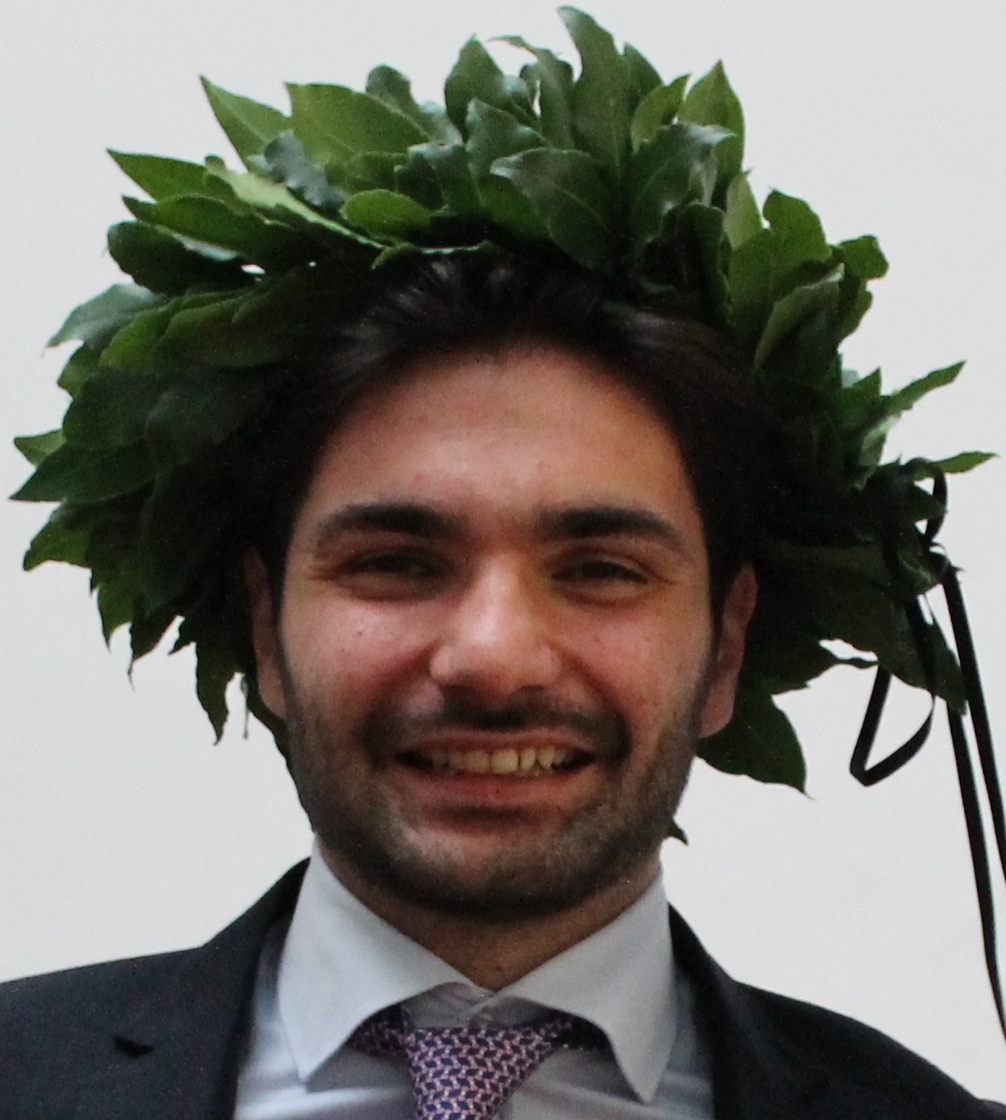}}]{Cosimo Rulli}
is a Ph.D. student at the University of Pisa and a researcher with the National Research Council of Italy. He received his Master's Degree from the University of Florence in 2019, with a thesis on deep neural network compression with knowledge distillation and pruning. His research interests focus on deep learning, model compression, and information retrieval.  
\end{IEEEbiography}

\begin{IEEEbiography}[{\includegraphics[width=1in,height=1.25in,clip,keepaspectratio]{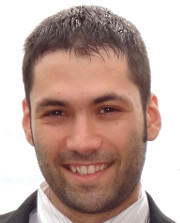}}]{Salvatore Trani}
is a researcher with the National Research Council of Italy. He received his Ph.D. in Computer Science from the University of Pisa in 2017. His main research interests range from Information Retrieval to Web Mining and Machine Learning. He authored more than 15 papers on these topics, published in peer-reviewed international journals and  conferences.
\end{IEEEbiography}

\begin{IEEEbiography}[{\includegraphics[width=1in,height=1.25in,clip,keepaspectratio]{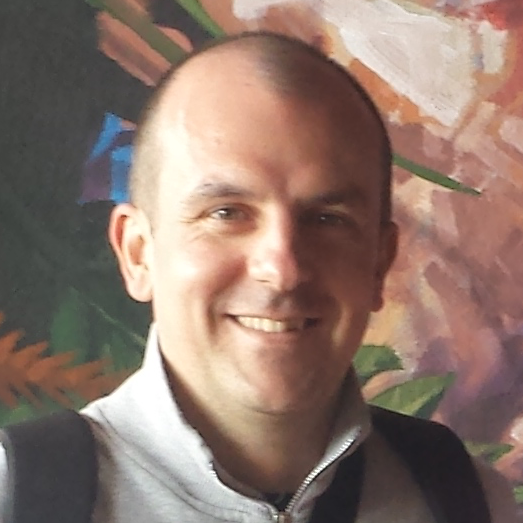}}]{Rossano Venturini}
received his Ph.D. degree from the University of Pisa, in 2010. He is an Associate Professor at the Computer Science Department of the University of Pisa. His research interests mainly focus on designing and analyzing algorithms and data structures for large datasets with applications in Information Retrieval and Machine Learning. He received two Best Paper Awards at ACM SIGIR in 2014 and 2015. For more information: \href{http://pages.di.unipi.it/rossano}{http://pages.di.unipi.it/rossano}.
\end{IEEEbiography}

\vfill

\end{document}